\documentclass[]{spie}  

\usepackage{amsmath,amsfonts,amssymb}
\usepackage{graphicx}
\usepackage{comment}
\usepackage{subcaption} 
\usepackage[colorlinks=true, allcolors=blue]{hyperref}

\title{Methods for evaluating the resolution of 3D data derived from satellite images}

\author{Christina Selby}
\author{Holden Bindl}
\author{Tyler Feldman}
\author{Andrew Skow}
\author{Nicolas Norena Acosta}
\author{Shea Hagstrom}
\author{Myron Brown}
\affil{The Johns Hopkins University Applied Physics Laboratory\\Laurel, Maryland, USA}

\begin{document} 
\maketitle

\begin{abstract}
3D data derived from satellite images is essential for scene modeling applications requiring large-scale coverage or involving locations not accessible by airborne lidar or cameras. Measuring the resolution of this data is important for determining mission utility and tracking improvements. In this work, we consider methods to evaluate the resolution of point clouds, digital surface models, and 3D mesh models. We describe 3D metric evaluation tools and workflows that enable automated evaluation based on high-resolution reference airborne lidar, and we present results of analyses with data of varying quality.
\end{abstract}


\section{INTRODUCTION}


Satellite images are an essential source of 3D data for large-scale scene modeling applications where it is impractical to collect with airborne lidar or cameras. 3D reconstruction pipelines for satellite images based on multi-view stereo (MVS) \cite{rsp_pipeline, s2p8014932, VisSat19, Leotta_2019_CVPR_Workshops} or neural radiance fields (NeRF) \cite{mari2022sat, Mari_2023_CVPR, zhang2024brdfnerfneuralradiancefields, sprintson2024fusionrfhighfidelitysatelliteneural} can produce 3D point clouds, digital surface models (DSM) and 3D mesh models with vertical accuracy that compare well with airborne lidar measurements. Reconstructing 3D shapes from multiple images with horizontal boundary precision is very challenging, particularly when objects in a scene are closely spaced. It is important to understand the limits of achievable horizontal resolution with these methods, both to characterize expected utility for practical applications and to track progress of research aimed at producing 3D higher-resolution products.

Prior work has been done to evaluate horizontal boundary precision for polygonal building models. For example, Bosch et al. \cite{isprs-archives-XLII-1-W1-239-2017} reported the root mean squared error (RMSE) for nearest horizontal distances between points sampled from predicted and reference building polygons, and Wang et al. \cite{Wang_2023_ICCV} reported average corner offsets (ACO) for polygon vertices. We are not aware of prior work to evaluate similar metrics for more general 3D data in the absence of semantic labels.

In our work, we seek to evaluate horizontal resolution for general 3D data derived from satellite images, including 3D point clouds, DSMs, and triangulated meshes. We draw inspiration from Stevens et al. \cite{2011_Stevens} who evaluated horizontal resolution of airborne lidar 3D point clouds by measuring the contrast transfer function (CTF) using intricately crafted test targets with tribars of varying widths and pair-wise distances. We propose a similar approach that leverages in-scene urban features. 

\section{METHODS}

\subsection{Contrast Transfer Function} \label{sec:ctf}
The contrast transfer function (CTF) has been defined for two-dimensional imagery as a way to characterize the spatial fidelity of an imaging system \cite{schott2007remote}.  The CTF is a measure of the discernible contrast of a tribar measure and is defined as 
\begin{equation}
    C = \frac{A-B}{A+B},\label{eq:ctf_basic}
\end{equation}
where $A$ and $B$ are the observed brightness for white and dark bars in a square-wave target.  The CTF can be analyzed as a function of distance between bars.  A CTF threshold is established (typically around $.1-.2$) for which it is assumed an observer can discern between targets.  The distance at which that threshold is met can be reported as the spatial resolution.  However, analysis of all of the CTF data is more complete than only reporting the distance at the selected CTF threshold. The CTF is approximately a measure of the edge spread function (ESF) under the assumption that the system response is the same in the $x$ and $y$ direction\cite{schott2007remote}.  

The CTF methodology presented here has been adapted in a way that is useful for the evaluation of provided 3D data products that measure elevation instead of image intensity.  Details of that adaptation are in the following section.  Sets of evaluation regions derived from parallel building footprints will be used as a proxy for a tribar measure. A local alignment of test and reference datasets will be described so that the intensities input into Equation \eqref{eq:ctf_basic} can be interpreted appropriately.  The analysis will provide a scatter plot of distance versus CTF values and variation in the calculated CTF will be observed due to causes such obstructions in the evaluation regions or changes in buildings between the reference and test data.  Further, there may be variation in the CTF signal due to the DSM creation methodology itself.  The function 
\begin{equation}
C(d) = A \ exp(-\left(\frac{\pi\sigma}{d}\right)^2), \label{eq:ctf_model}
\end{equation}
will be used to model the CTF data where $d$ is the distance between two targets and $\sigma$ and $A$ are fitted parameters.  This model will be used for determining a distance value for a given CTF threshold.  The motivation for the model is that the Fourier transform of a point-spread function is closely related to the CTF\cite{schott2007remote}.  If it is assumed that a point-spread function for the 3D data derivation process is Gaussian, the Fourier transform is $exp\left(-\left(\frac{\pi\sigma}{d}\right)^2\right)$.  The expected value of the CTF is not expected to be $1$ as $d$ approaches infinity because the test data is not assumed to match the reference data (regardless of distance between targets).  For this reason, the constant parameter $A$ is used to model this limit.

\begin{figure}[ht]
    \begin{center}
        \begin{tabular}{cc}
            \subcaptionbox{$E(CTF) = .5\, \mathrm{m}$ \label{fig:down1}}{
                \includegraphics[height=4.2cm]{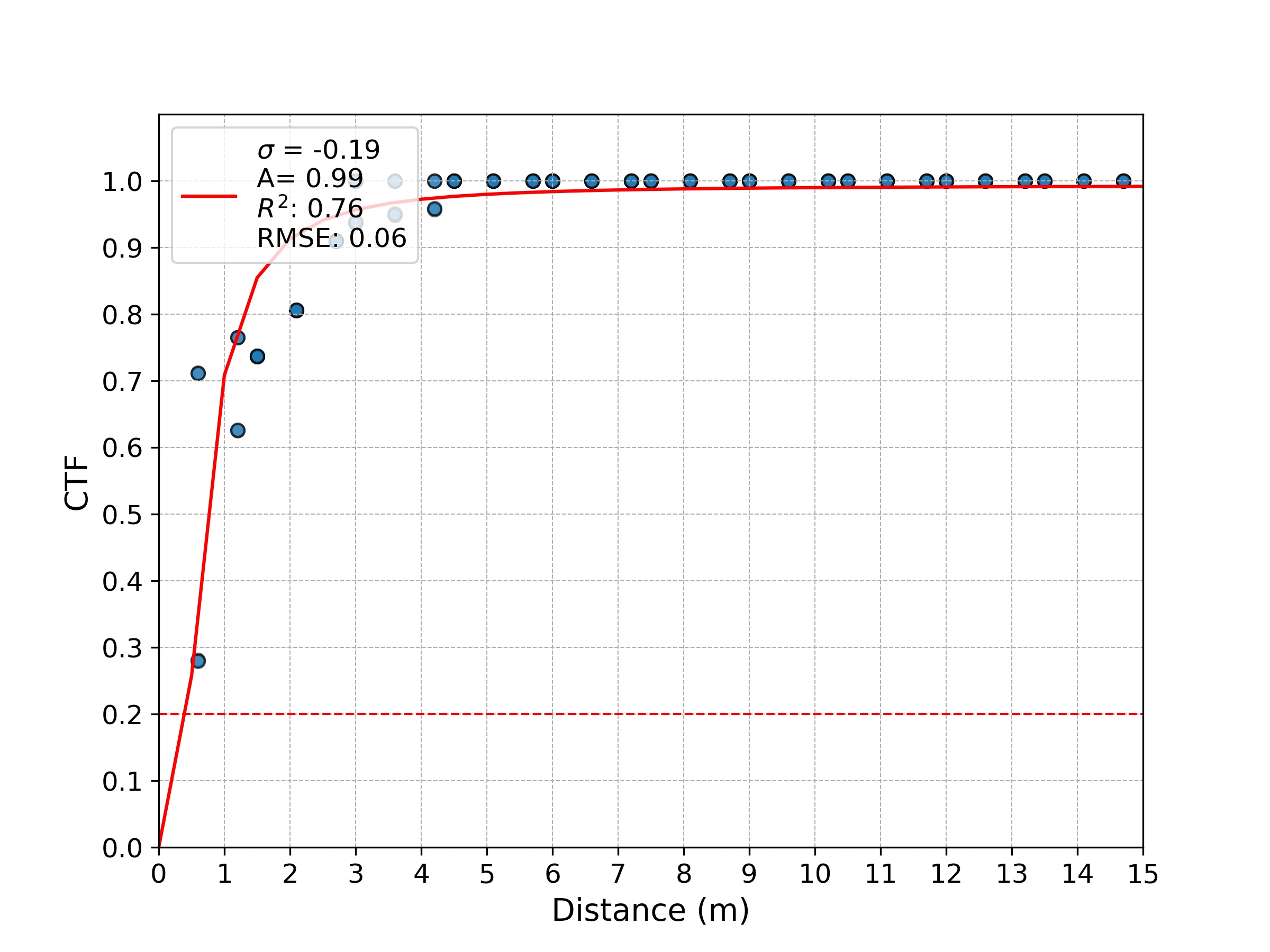}
            } &
            \subcaptionbox{$E(CTF)=1\, \mathrm{m}$\label{fig:down2}}{
                \includegraphics[height=4.2cm]{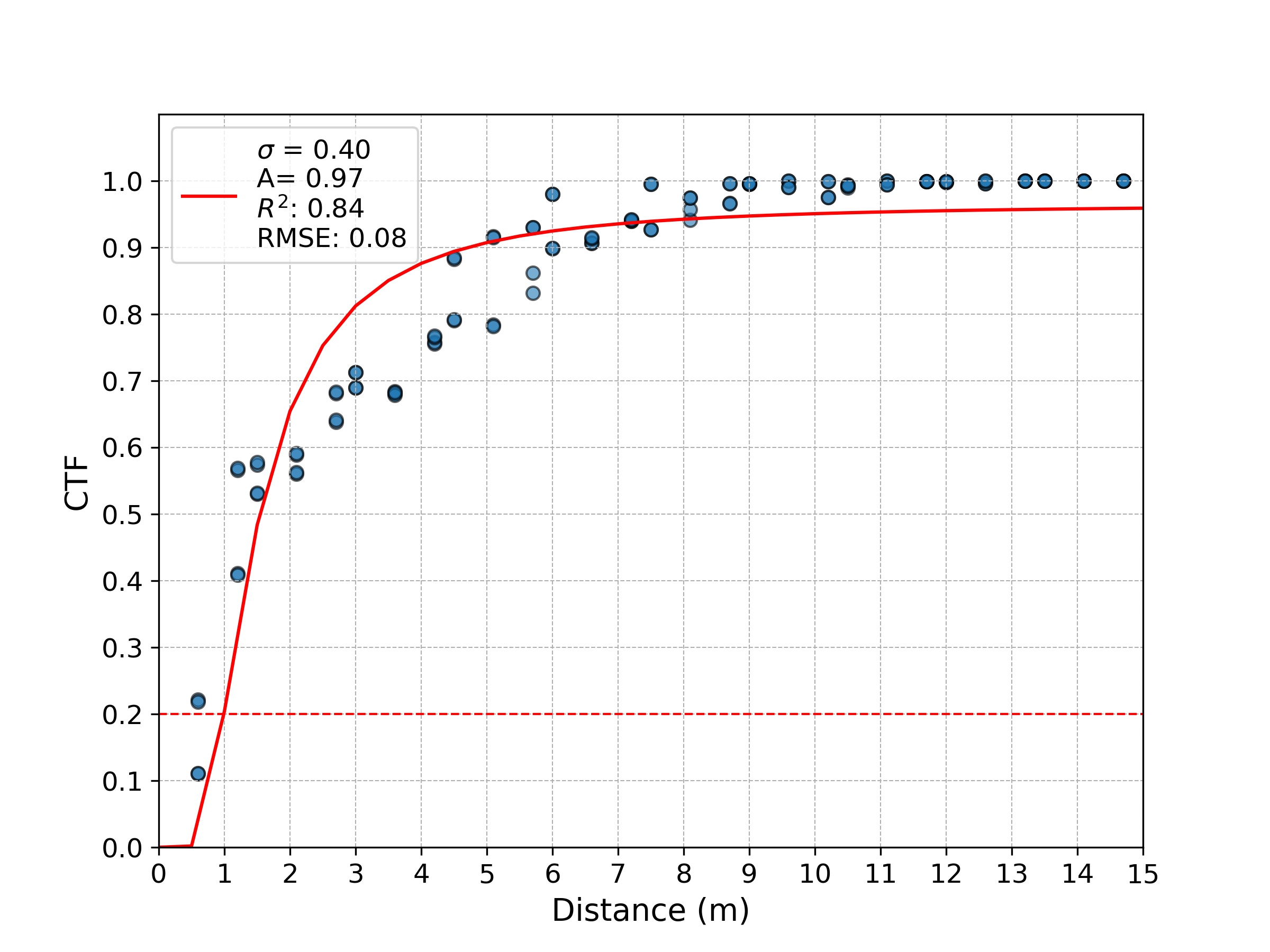}
            } \\
            \subcaptionbox{$E(CTF) = 2\, \mathrm{m}$\label{fig:down3}}{
                \includegraphics[height=4.2cm]{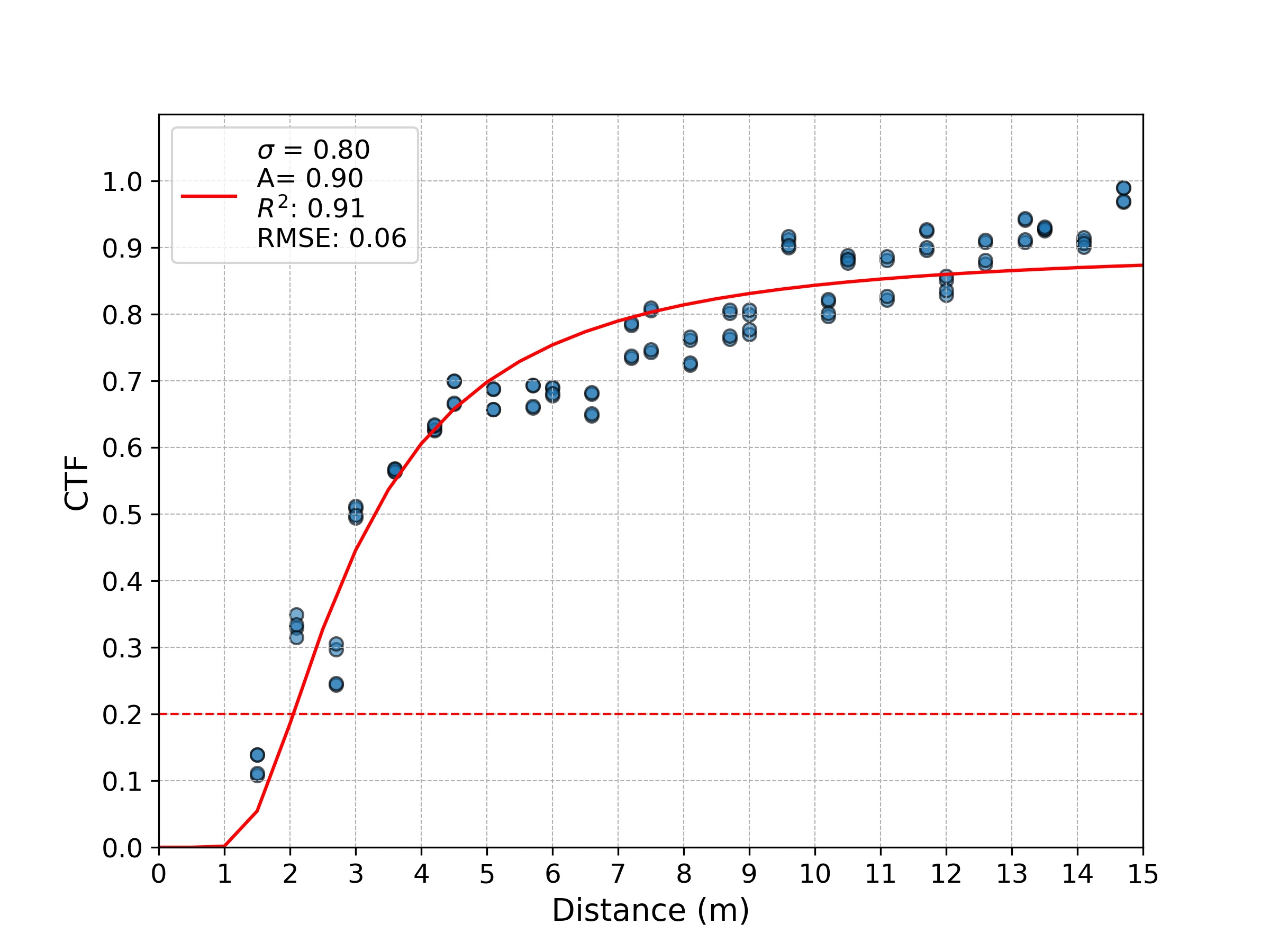}
            } &
            \subcaptionbox{$E(CTF)=4\, \mathrm{m}$\label{fig:down4}}{
                \includegraphics[height=4.2cm]{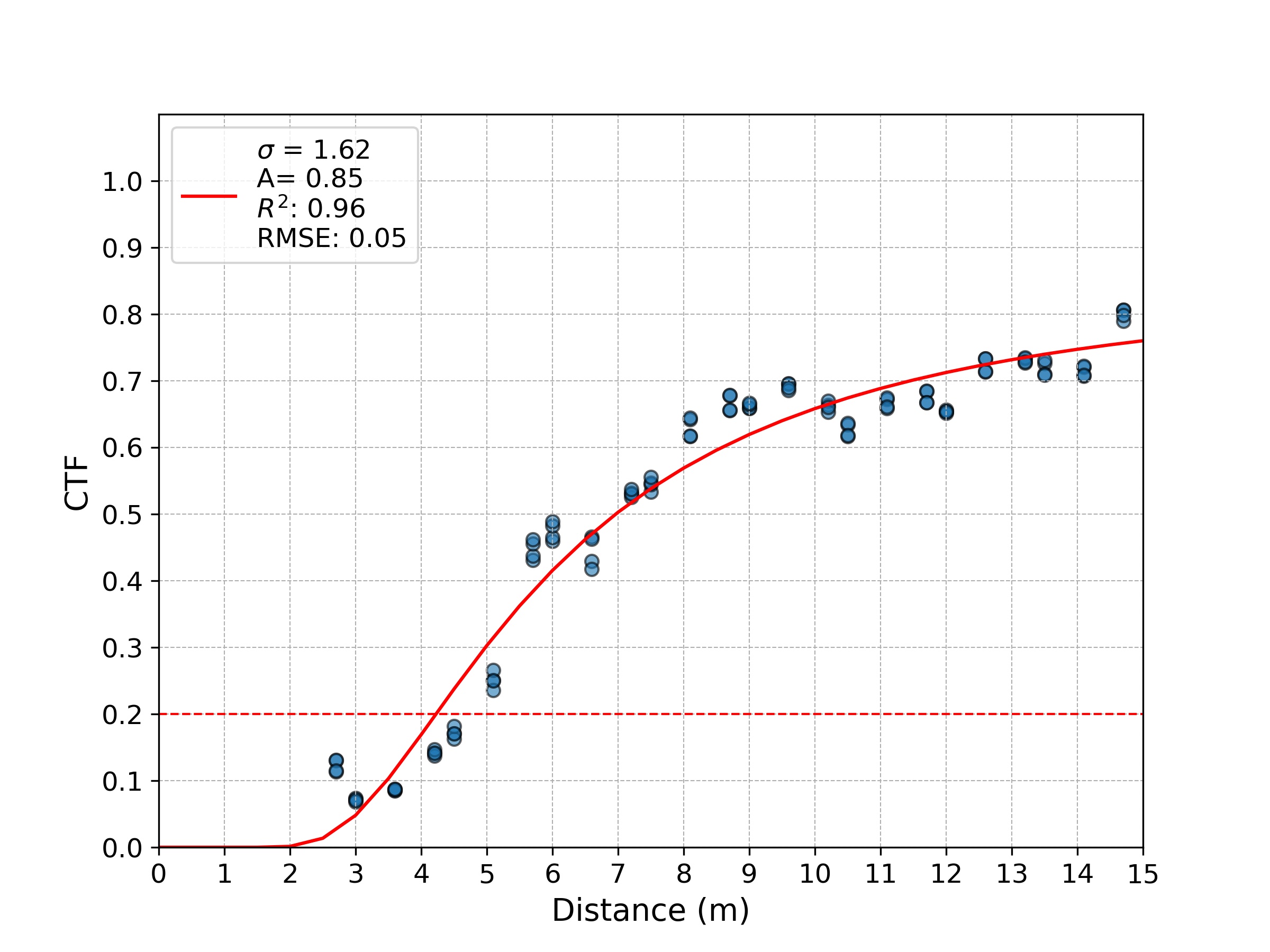}
            }
        \end{tabular}
    \end{center}
    \caption[example]{
        A tribar reference dataset was created for analysis of the CTF methodology. Figures (a) - (d) are CTF summary plots obtained after downsampling the test data by a factor of 2, 4, 8, 16.  The test data data was generated with a ground sample distance of $.25$ meters.  The expected distance values at a CTF value of $.2$ are then approximately $.5$, $1$, $2$, and $4 \, \mathrm{m}$, respectively.  The model fit agrees with these expected values.
    }
    \label{fig:ctf_model_fits}
\end{figure}

In order to demonstrate that Equation \eqref{eq:ctf_model} is a reasonable model choice, a set of data was created for which the distance value at which $C = .2$ is known.  In particular, a DSM modeling a tribar measure was created and  progressively downsampled by two.  The ground sample distance for the original DSM was $.25\, \mathrm{m}$ and so the assumed value of $d$ at $C=.2$ was $d = .25\, \mathrm{m}$.  The downsampled DSMs then had expected CTF values of $C=.2$ at $d = 2 \times .25 = .50\, \mathrm{m}$, $d = 4 \times .25 = 1\, \mathrm{m}$, $d = 8 \times .25 = 2 \, \mathrm{m}$, and $d = 16 \times .25 = 4\, \mathrm{m}$ with each consecutive downsampling.  The CTF data and the fitted models are shown in Figure \ref{fig:ctf_model_fits}.  

The interpretation of CTF for a provided 3D data product is the same as interpreting the performance of an imaging system.  One may interpret the 3D CTF calculation as an evaluation of the performance of a \emph{3D reconstruction process}, which could include algorithms.  Further, the performance of relevant algorithms could depend on characteristics of the data that was used as input to these algorithms.  For this reason, care should be taken when analyzing the CTF results for a given 3D product.  The quality of the 3D data reconstruction may depend not just on the methodology but also on imagery inputs and the elevation characteristics of the region. The CTF evaluation described here could be useful for comparing multiple data reconstruction processes by considering the outputs of these processes over a significant set of sites.  The CTF evaluation described here could also be useful for comparing the outputs of the same 3D data reconstruction process utilizing different sets of inputs.

\subsection{CTF Data Pipeline}
This section will describe the details in the steps necessary for estimating the CTF function from a given test and reference 3D dataset.  Visualizations of the steps for an example over Nellis Air Force Base are provided.  The input to the CTF data pipeline is a reference lidar point cloud and a 3D test data set which can be a point cloud, digital surface model, or 3D mesh model.  Evaluation regions are obtained from parallel building footprint pairs and a CTF value calculated for each evaluation region.  Building footprints can be derived from a segmentation model applied to the reference lidar point-cloud.  Other building footprint sources, such as Open Street Maps (OSM) \cite{OpenStreetMap}, can also be utilized after being aligned to the reference dataset. The pipeline produces GeoJSON files with the evaluation region geometries attributed with the CTF values.  These GeoJSON files can be loaded into a GIS viewer for analysis.  Summary plots are also created that shows the relationship between distance and CTF and the fitted model given by Equation \eqref{eq:ctf_model}.

\subsubsection{Data Preparation}
\begin{figure}[ht]
    \begin{center}
        \begin{tabular}{cccc}
            \subcaptionbox{Point Cloud\label{fig:nellis_points}}{
                \includegraphics[height=3.3cm]{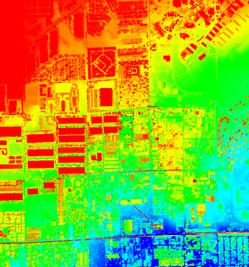}
            } &
            \subcaptionbox{Class Labels\label{fig:nellis_with_classes}}{
                \includegraphics[height=3.3cm]{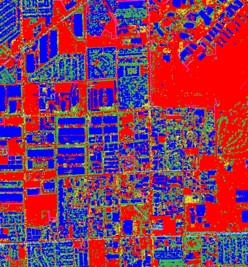}
            } &
            \subcaptionbox{DSM\label{fig:nellis_dsm}}{
                \includegraphics[height=3.3cm]{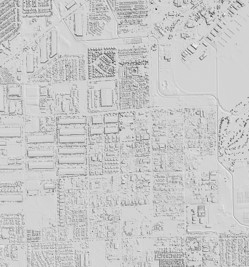}
            } &
            \subcaptionbox{DTM\label{fig:nellis_dtm}}{
                \includegraphics[height=3.3cm]{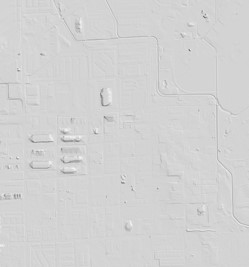}
            }
        \end{tabular}
    \end{center}
    \caption[example]{
         This example from the Nellis test site illustrates the 3DEP lidar  point cloud \cite{USGS_3DEP}, class labels predicted for each point, derived DSM, and derived DTM. The class labels, DSM, and DTM are derived from the 3DEP lidar point cloud\cite{USGS_3DEP}.  Any errors in class labels, most commonly observed for portions of larger buildings, are visually apparent in the derived DTM and result in inaccurate polygon extraction for those buildings. 
    }
    \label{fig:pointcloud_dsm}
\end{figure}

A Digital Surface Model (DSM) is produced from the reference point cloud by first estimating the Average Nominal Point Spacing (ANPS) to define the Ground Sample Distance (GSD) of the DSM grid and then mapping 3D point cloud values to gridded coordinates. To avoid aliasing, each point is evaluated at plus and minus one half GSD. The maximum Z value is assigned to the grid coordinate. No smoothing is applied. Any points labeled withheld in the reference point cloud are excluded. A Digital Terrain Model (DTM) is also produced by assigning minimum Z values, including only points labeled ground in classification, and interpolating to fill any unassigned grid coordinates. An example of derived reference data is shown in Figure \ref{fig:pointcloud_dsm}.

The test data is converted to the same coordinate reference system as the reference data.  The test data is projected so that it uses the same affine transform to map geospatial coordinates to pixel coordinates as the reference data.  The test and reference data is assumed to be similar enough that one could observe if the data is aligned.  The test data is aligned horizontally and vertically to the reference data by considering a set of subwindows over the data.  If the subwindow contains more than ninety-five percent valid test data then a phase correlation algorithm is applied to calculate an $x$ and $y$ correction.  A vertical correction is determined by calculating the median difference of the test and reference elevation values.  The median of each correction is calculated over the subwindows and used to transform the test data. The test to reference alignment results for this presented Nellis example yielded an x-offset of $-1.187\, \mathrm{m}$, y-offset of $0.547\, \mathrm{m}$, and z-offset of $0.005\, \mathrm{m}$. 

\subsubsection{Building Footprint Generation}
Building footprints are essential to the evaluation of CTF and take the place of a tribar measure.  The pipeline can utilize as direct input any building footprints stored as a GeoJSON file.  The pipeline also has the ability to generate a GeoJSON file of building footprints from Open Street Maps or derive footprints directly from the outputs of a lidar segmentation model.

A point cloud classification algorithm is applied to the reference lidar point cloud to assign labels to each point for ground, vegetation, building, wall, power line, civilian vehicle, truck, military vehicle, aircraft, and pole. The algorithm uses an SPUNet\cite{liu2022spunetselfsupervisedpointcloud} model implemented in the Pointcept\cite{pointcept2023} framework and trained using  DALES\cite{varney2020dales}, STPLS3D\cite{Chen_2022_BMVC}, and swissSURFACE3D\cite{swissSURFACE3D} datasets.

The output of the lidar segmentation model is used to create a binary building mask.  This is completed by thresholding the confidence outputs of the model and mapping the location of a point in the point cloud to a pixel location and assigning a pixel value of $1$.    Geometries are extracted from the binary building mask and used as input to the Douglas-Peucker algorithm \cite{douglas1973algorithms} for simplification. This method has been seen to be sufficient for the CTF use case, but further work could be completed to improve the quality of the building footprints such as using frame fields \cite{Girard_2021_CVPR}. A huge benefit of using the reference lidar derived footprints is that they are aligned to the reference data by default.  A risk in using the reference lidar derived footprints is that sometimes transient structures are detected that are not present in the test dataset.  

The lidar segmentation model output is also used for aligning other building footprints to the reference data.  For each provided footprint, a translation is determined which minimizes the difference in the ground mask and the building mask over the footprint. 

\begin{figure}[ht]
    \begin{center}
        \begin{tabular}{ccc} 
            \subcaptionbox{Reference DSM\label{fig:osm_alignment_a}}{
                \includegraphics[height=3.5cm]{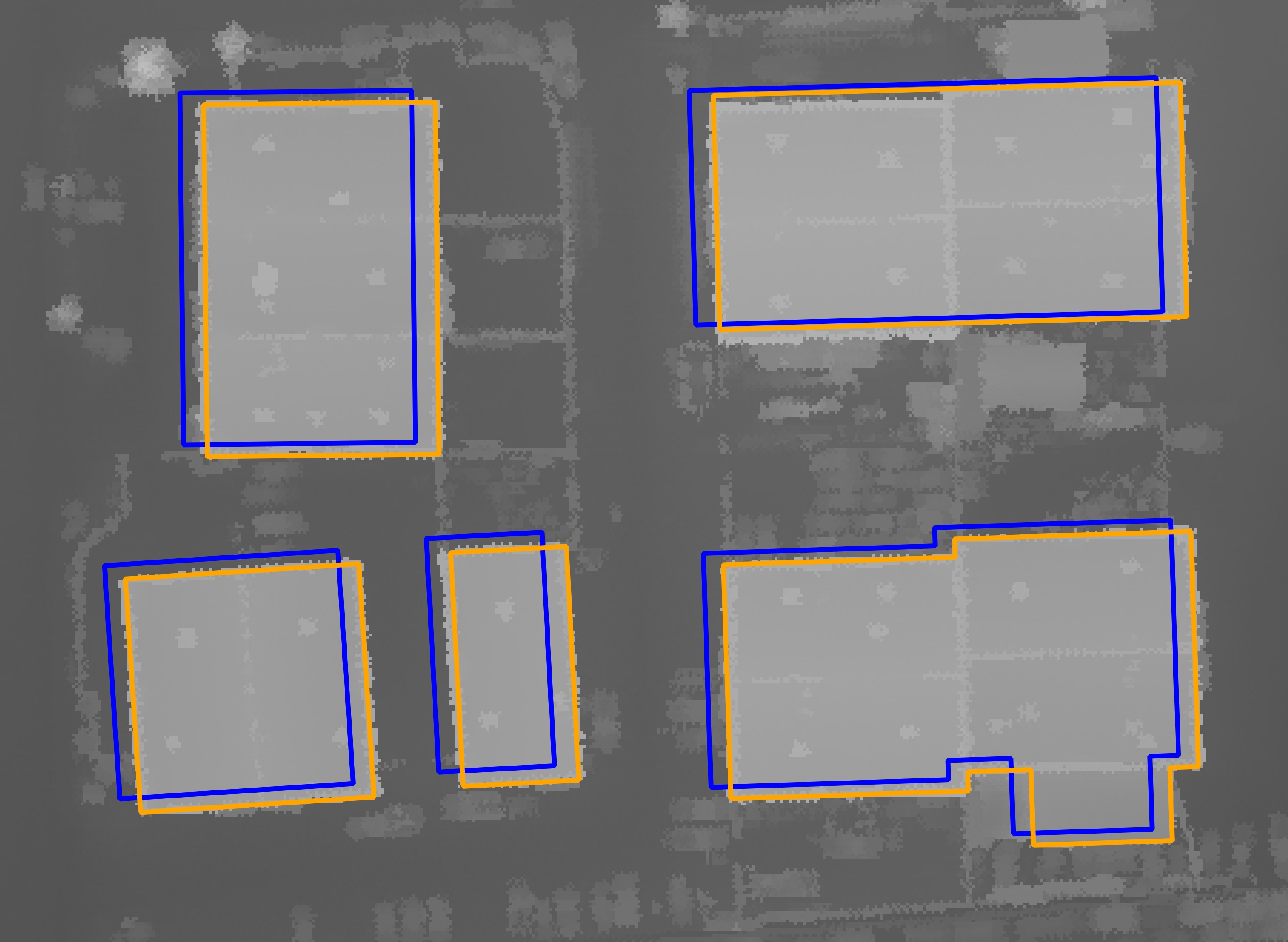}
            } &
            \subcaptionbox{Building Mask\label{fig:osm_alignment_b}}{
                \includegraphics[height=3.5cm]{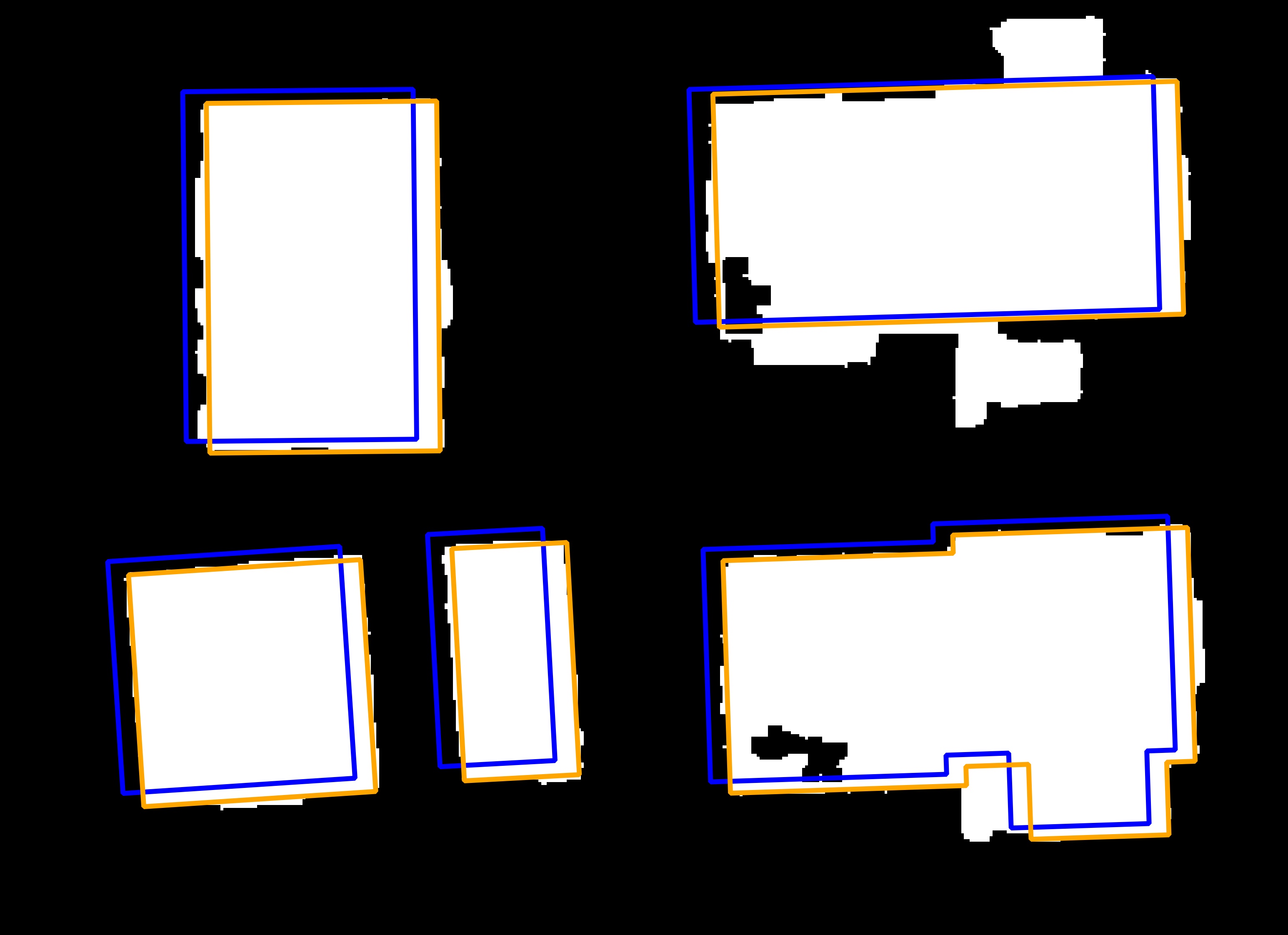}
            } &
            \subcaptionbox{Ground Mask\label{fig:osm_alignment_c}}{
                \includegraphics[height=3.5cm]{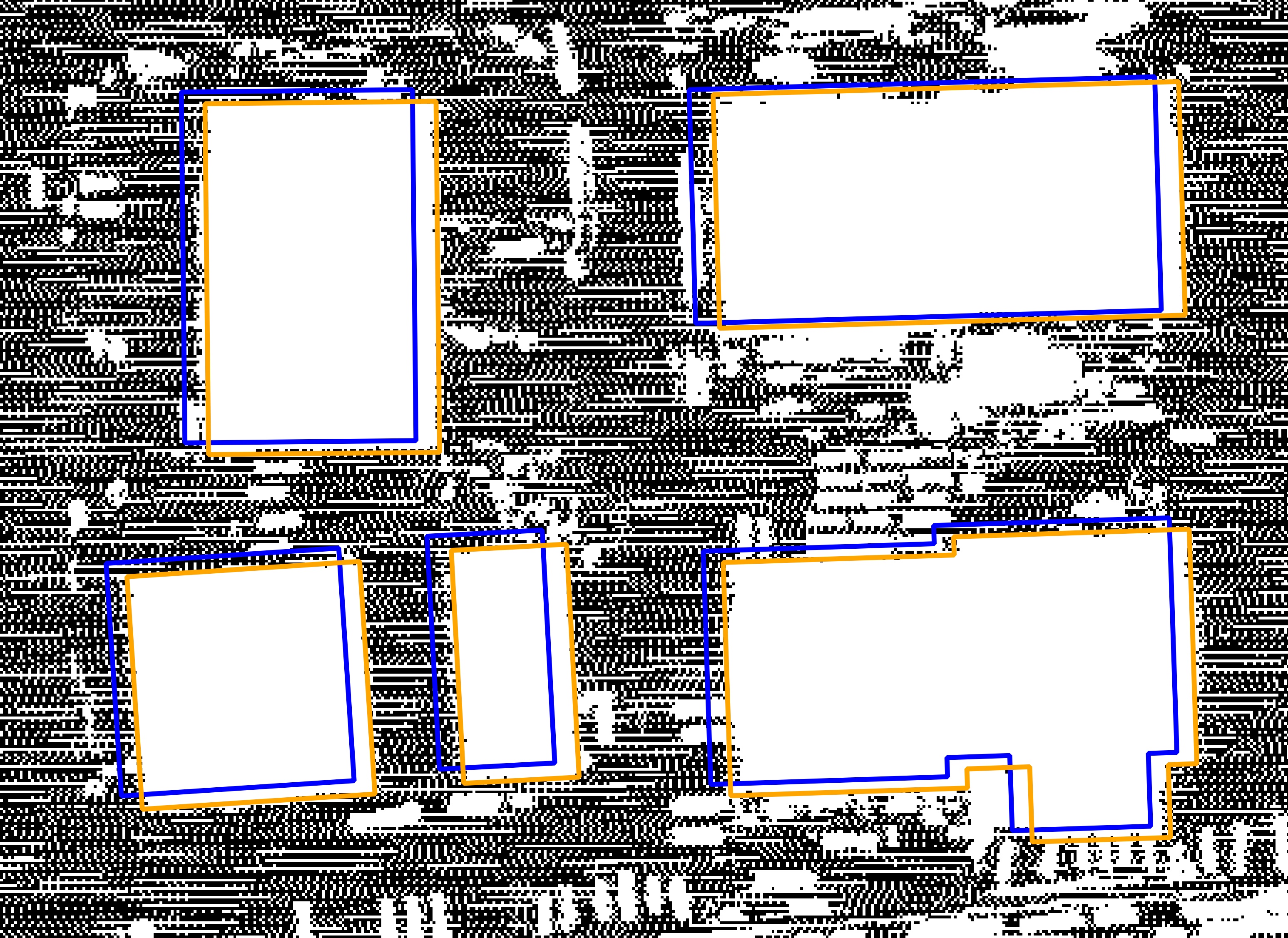}
            }
        \end{tabular}
    \end{center}
    \caption[example]{
        Original OSM footprints\cite{OpenStreetMap} (blue) and aligned OSM footprints (orange) over the reference digital surface model, the building mask derived from the lidar segmentation model, and the derived ground mask.  Note that the ground mask has not had any morphological operations applied, while the building mask has.  The  reference DSM, building mask, and ground mask were derived from the 3DEP lidar  point cloud \cite{USGS_3DEP} over a small region near Nellis AFB.
    }
    \label{fig:osm_alignment}
\end{figure}

\begin{figure}[ht]
    \begin{center}
        \begin{tabular}{cc} 
            \subcaptionbox{Reference DSM\label{fig:segmentation_comparison_a}}{
                \includegraphics[height=4cm]{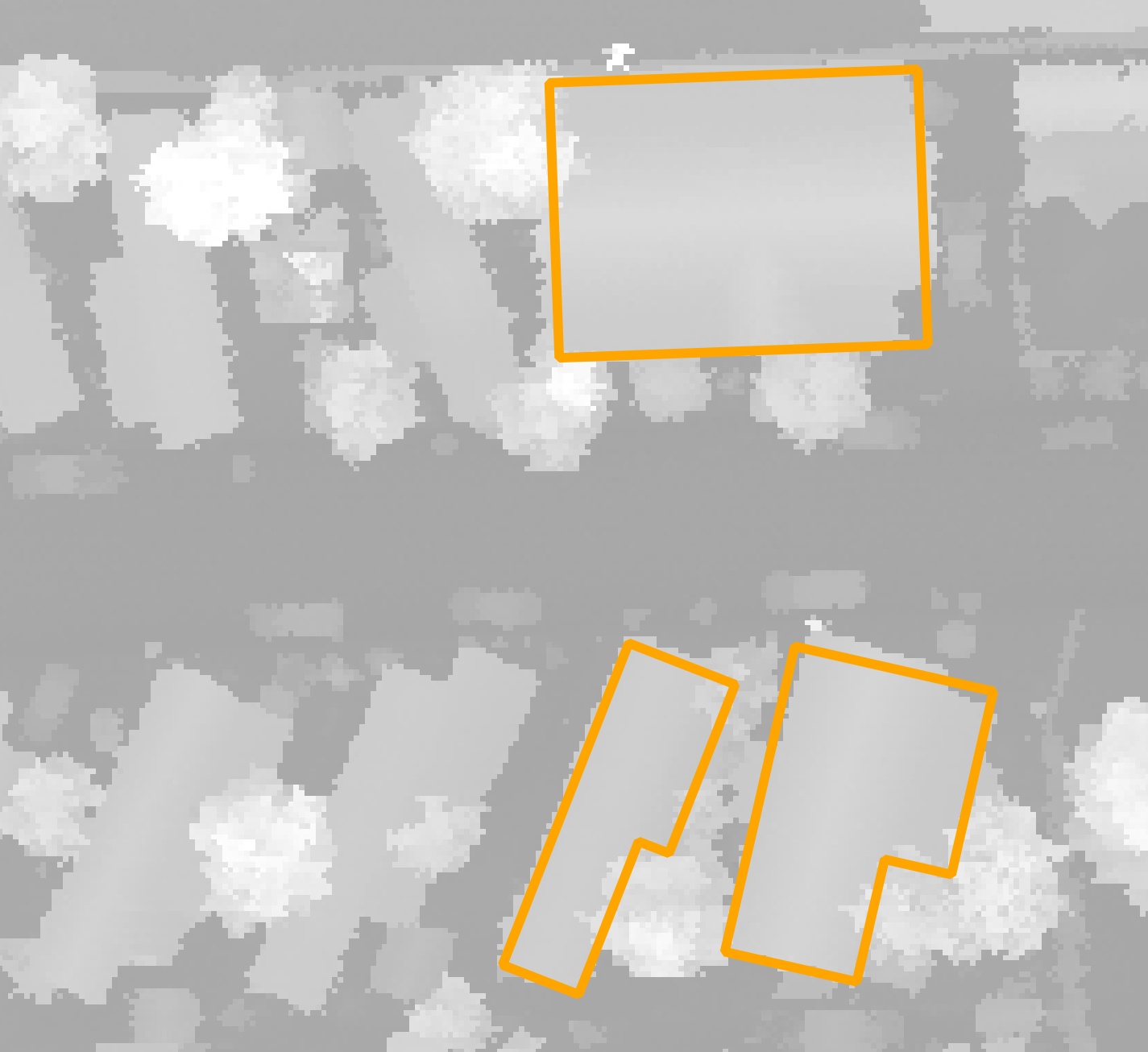}
            } &
            \subcaptionbox{Building Mask\label{fig:segmentation_comparison_b}}{
                \includegraphics[height=4cm]{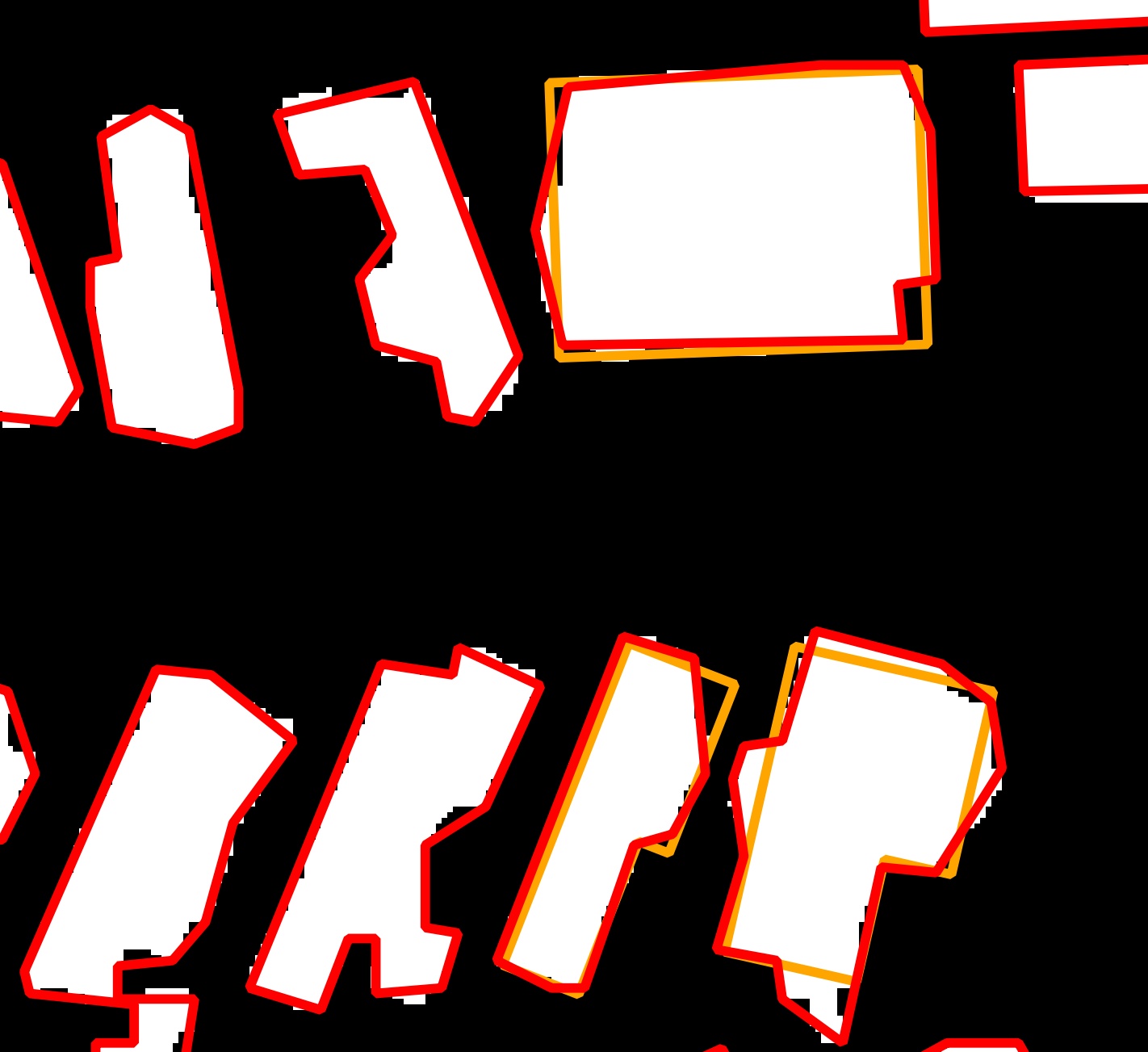}
            }
        \end{tabular}
    \end{center}
    \caption[example]{
        Aligned OSM footprints\cite{OpenStreetMap} (orange) and segmentation model derived footprints (red) over the derived reference DSM and building mask derived from the lidar segmentation model. The  reference DSM and building mask were derived from the 3DEP lidar  point cloud \cite{USGS_3DEP} over a small region near Nellis AFB.
    }
    \label{fig:segmentation_comparison} 
\end{figure}

An example of the original and aligned OSM footprint is in Figure \ref{fig:osm_alignment} along with the relevant segmentation model derived masks.  A comparison of the OSM building footprints and segmentation model derived footprints are provided in Figure \ref{fig:segmentation_comparison}.

\subsubsection{Evaluation Regions}
An evaluation region consist of three polygons; a center polygon over the ground and two parallel rectangles adjacent to the center region over buildings.  Parallel buildings take the place of a tribar measure in the CTF calculation and the building footprints are used as input for determining evaluation regions.  The centroid of each building is determined and the distance between each pair of centroids is calculated so building pairs can filtered to be less than a maximum centroid distance.  The Cartesian distance is then calculated between each remaining building pair and the building pairs are further filtered to include building pairs with orthogonal distance up to a provided maximum distance. This set of building pairs is then used to find evaluation regions.  

Given a pair of building footprints, an attempt is made to define an evaluation region through processing the building edges. The edges of the polygons are extracted from each building and a set of edge pairs is created (one edge from each building).  The edge pairs are filtered so that they are approximately parallel and the region between them does not intersect other edges.  The center region is defined by creating a rectangle from the edge pairs.  A region over each building is then determined which is approximately the same size as the center region and adjacent to the center region. Only one evaluation region is defined for each building pair; the one with the smallest orthogonal distance between the buildings.  Future analysis could consider including all possible evaluation regions instead of just the one with smallest orthogonal distance.  Improvements to the segmentation model footprints could also lead to finding more evaluation regions.

\begin{figure}[ht]
    \begin{center}
        \begin{tabular}{cc} 
            \subcaptionbox{Reference DSM\label{fig:ref_region}}{
                \includegraphics[height=3.75cm]{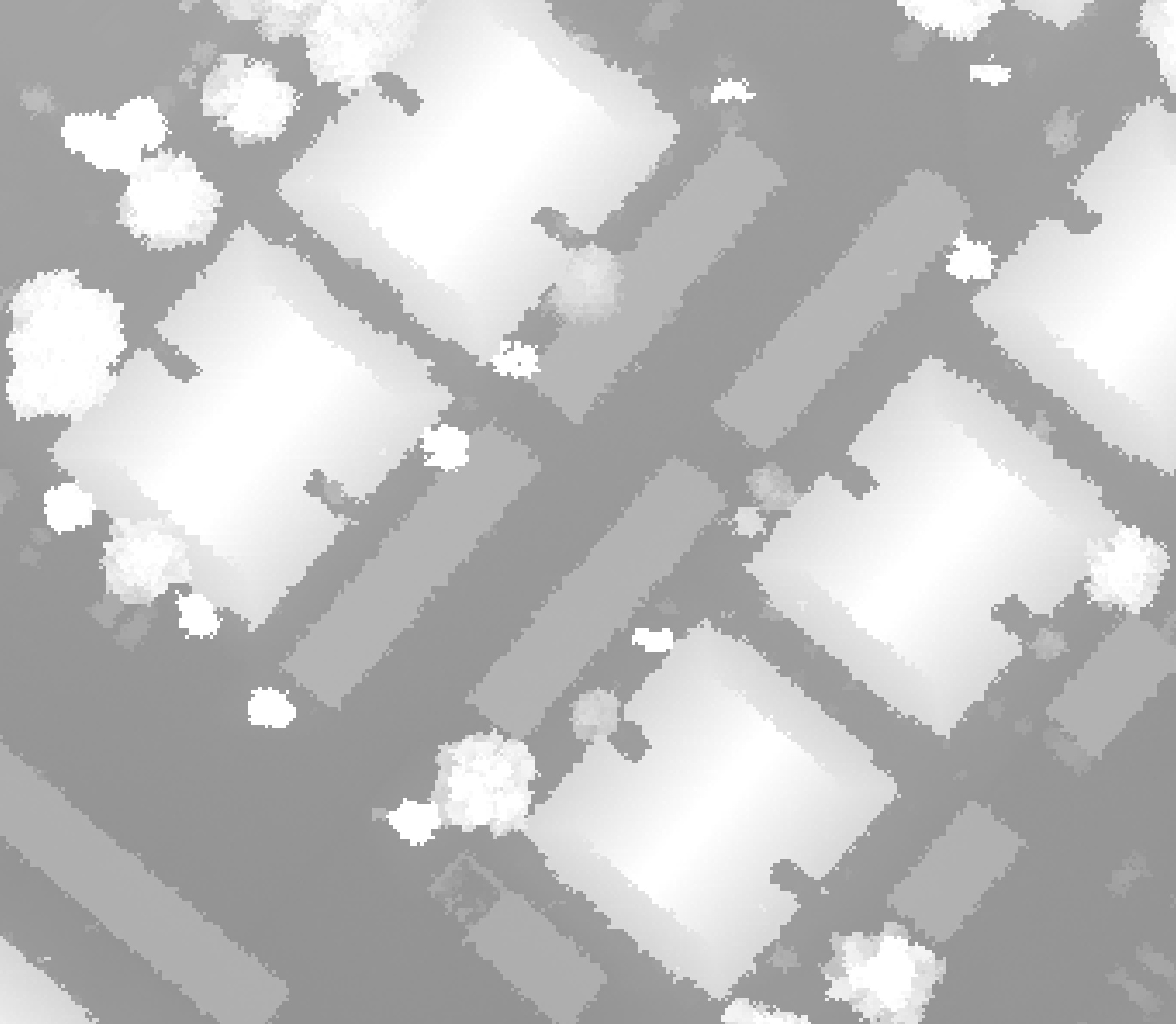}
            } &
            \subcaptionbox{Test DSM\label{fig:test_region}}{
                \includegraphics[height=3.75cm]{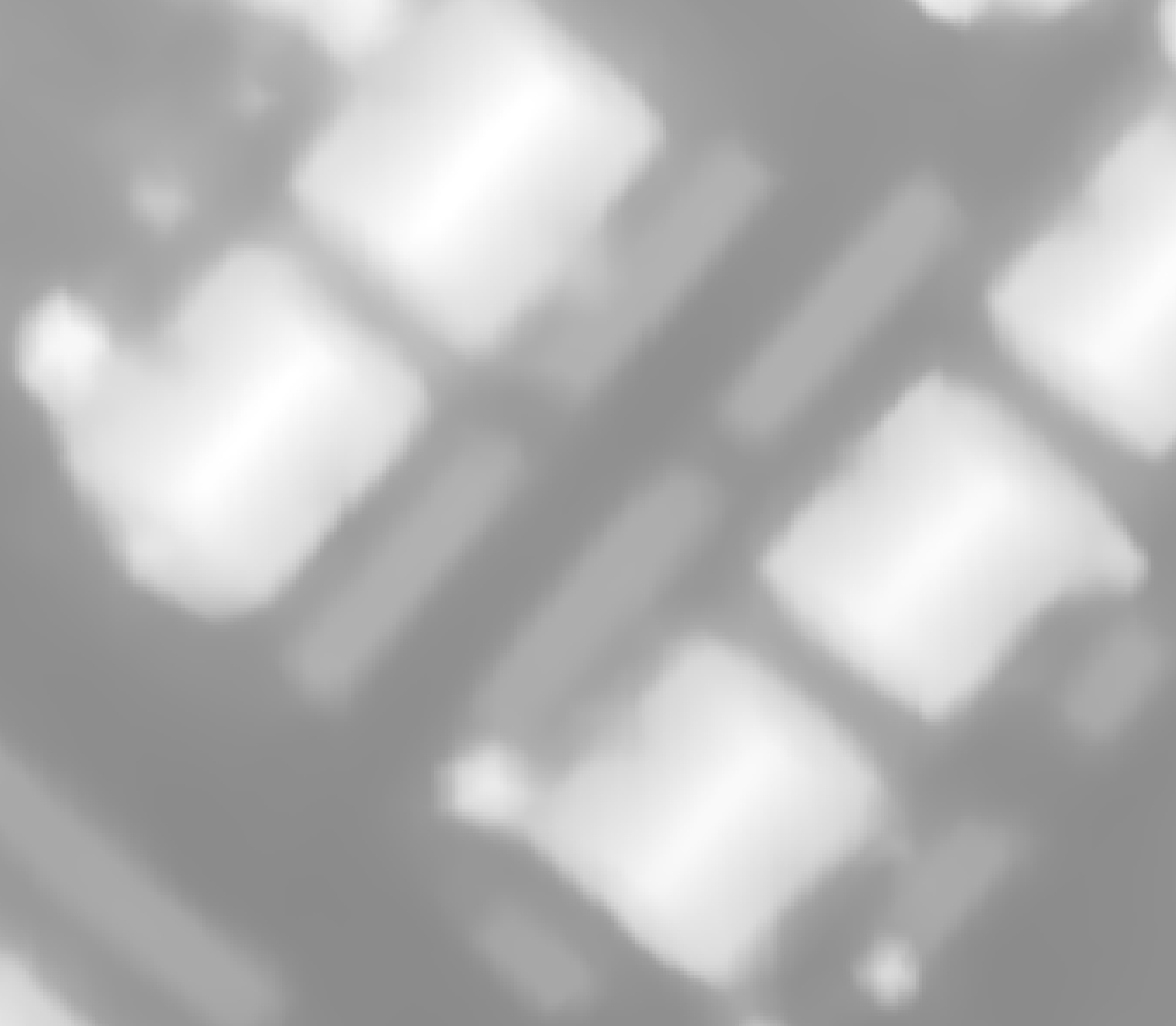}
            } \\
            \subcaptionbox{Evaluation Regions over Reference\label{fig:ref_eval_reg}}{
                \includegraphics[height=3.75cm]{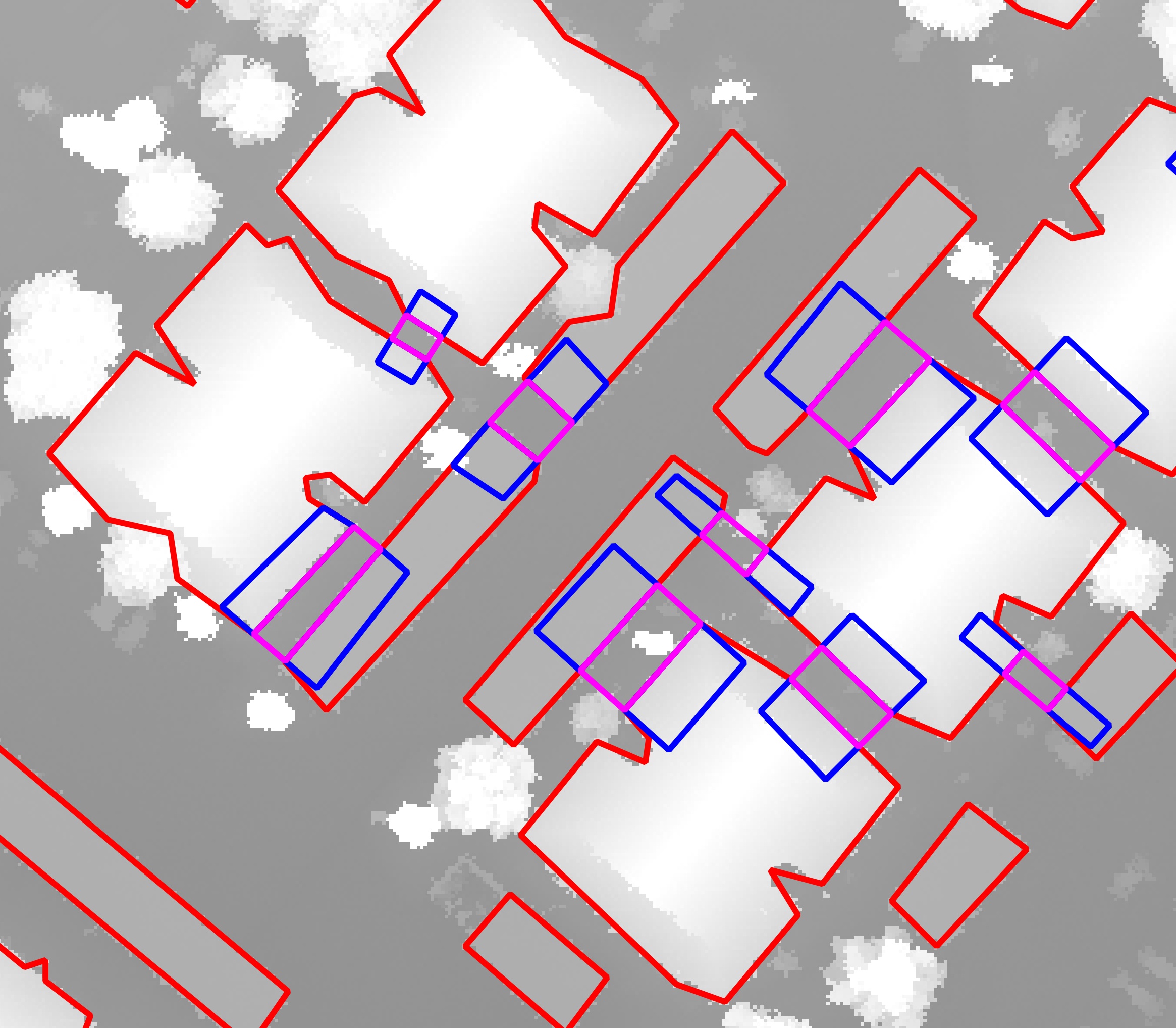}
            }& 
            \subcaptionbox{Evaluation Regions over Test\label{fig:test_eval_reg}}{
                \includegraphics[height=3.75cm]{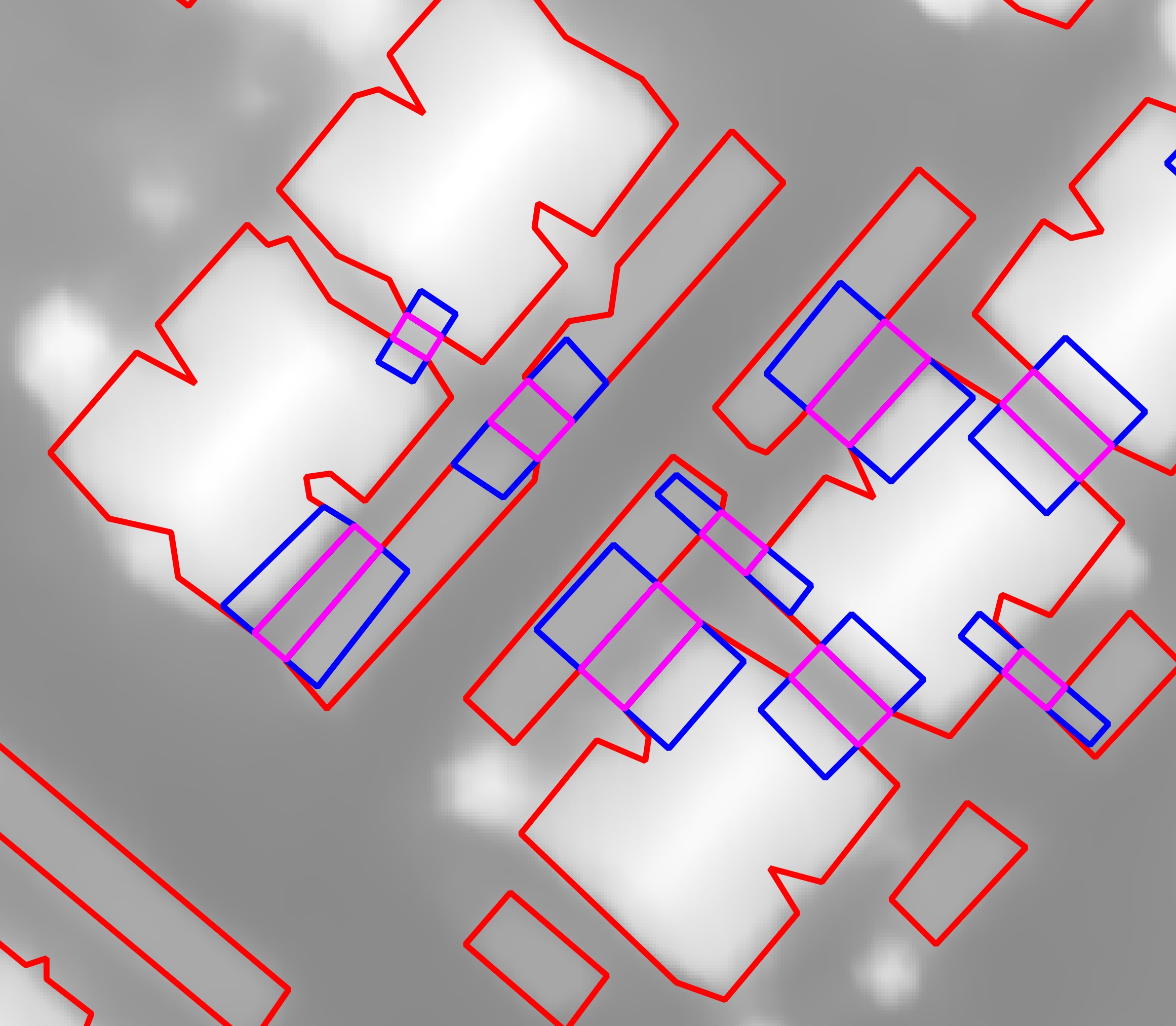}
            }
        \end{tabular}
    \end{center}
    \caption{Example subregion of reference (a) and test (b) DSM over Nellis.  Evaluation regions over the derived reference (c) and test (d) DSMs. Red denotes the lidar segmentation model derived footprints, blue the evaluation regions over the buildings and pink the center evaluation regions.  The reference DSM was derived from the 3DEP lidar  point cloud \cite{USGS_3DEP} over a small region near Nellis AFB and the test DSM was derived from MAXAR\textsuperscript{\copyright} satellite imagery, USGP-20211007}
    \label{fig:test_ref_compare}
\end{figure}

An example subregion of the reference and test DSMs and the overlaid corresponding evaluation regions are given in Figure \ref{fig:test_ref_compare}.

\subsubsection{CTF Calculation} \label{sec:ctf_calc}
A CTF value is calculated for each evaluation region.  The test and reference data represent elevation but using elevation values in Equation \eqref{eq:ctf_basic} gives a CTF value close to zero since the change in elevation with respect to elevation is very small.  A more appropriate zero value must be defined for the CTF calculation to be meaningful.  The tenth percentile of the reference data over the center region is used to define the elevation that is mapped to zero.

A local alignment over the building pair evaluation region is completed by centering the test data between the min and max of the reference data over the evaluation region. The motivation behind this alignment method is to consider how test data would appear if it were a Gaussian smoothed version of the reference data and to ensure consistency of alignment across the evaluation regions.  

\begin{figure}[ht]
    \begin{center}
        \begin{tabular}{ccc} 
            \subcaptionbox{\label{fig:loc_alignment_a}}{
                \includegraphics[height=3.75cm]{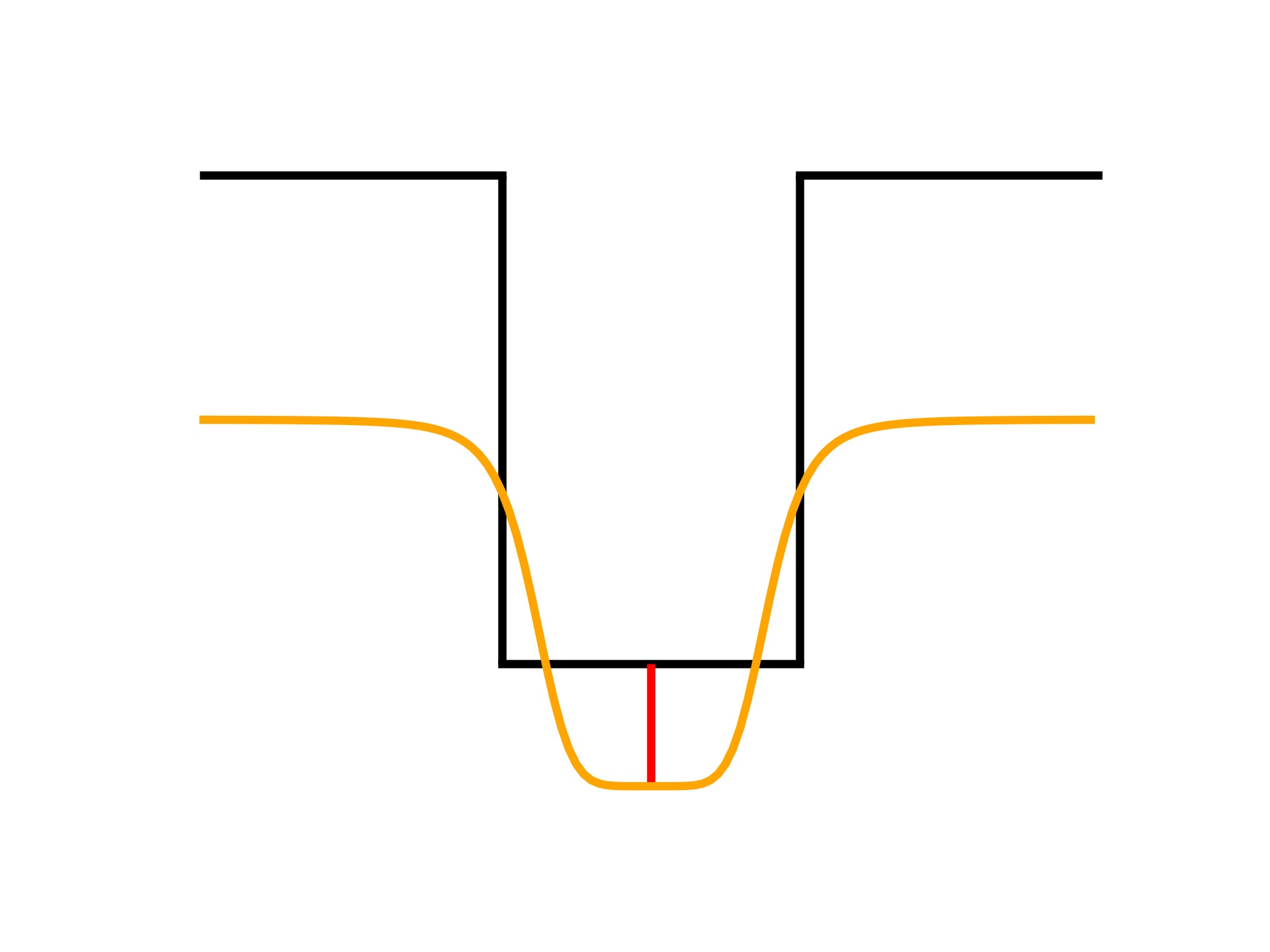}
            } &
            \subcaptionbox{\label{fig:loc_alignment_b}}{
                \includegraphics[height=3.75cm]{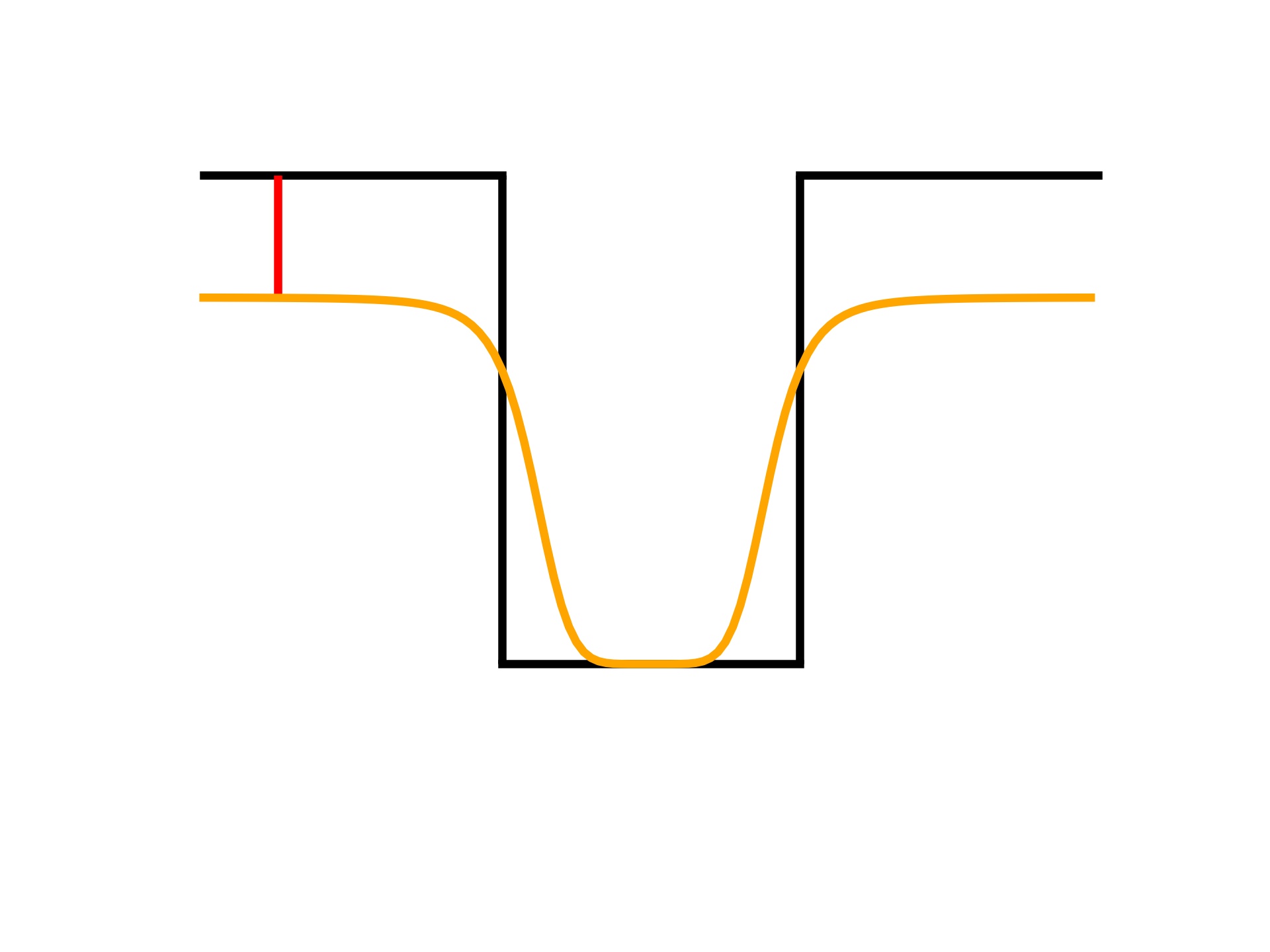}
            } &
            \subcaptionbox{\label{fig:loc_alignment_c}}{
                \includegraphics[height=3.75cm]{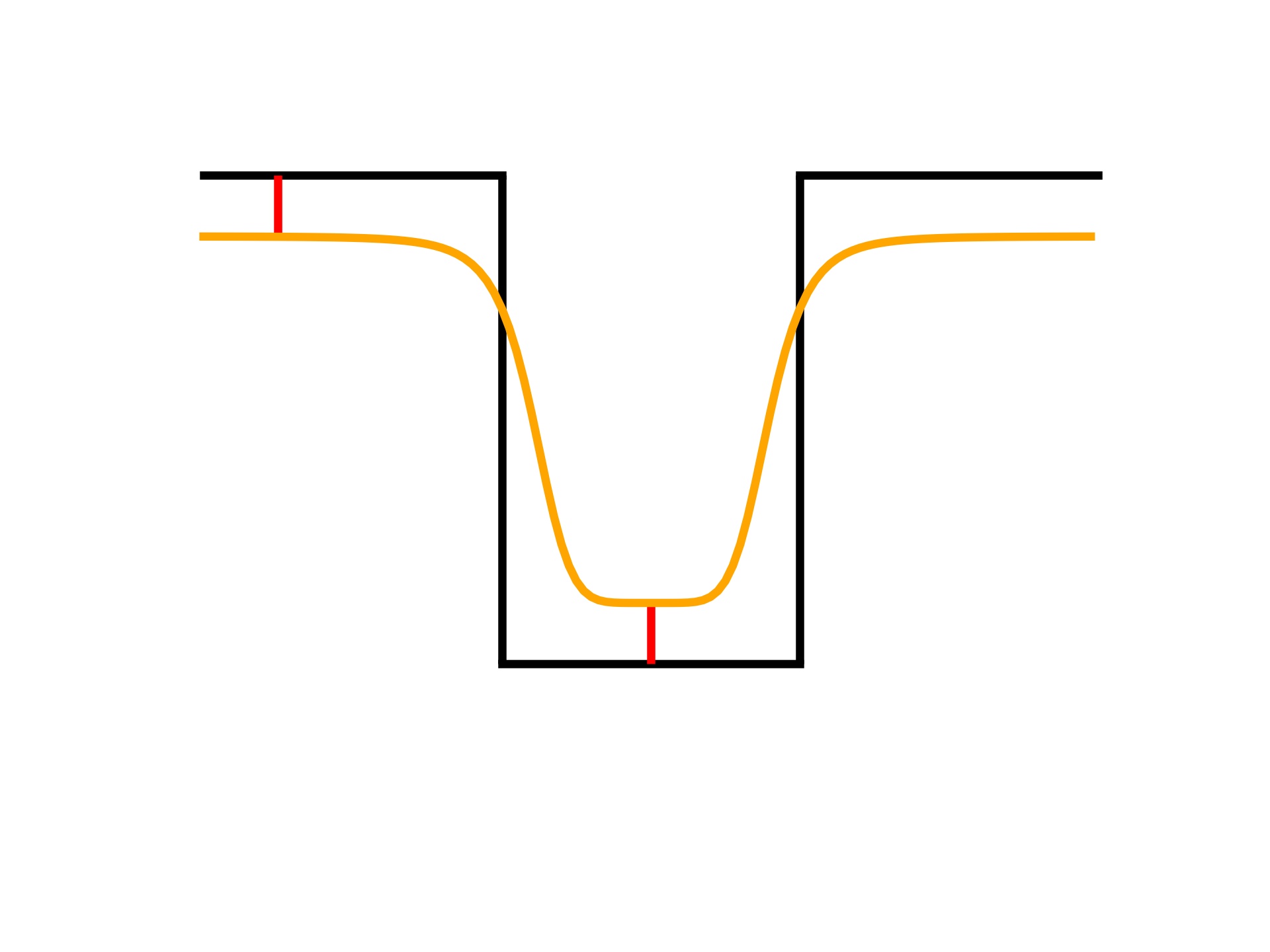}
            }
        \end{tabular}
    \end{center}
    \caption[example]{
        Figures (a)-(c) demonstrate the steps of the local alignment process.  In all figures black represents the reference data and orange the test data. In (a) an offset between the test and reference over the center region is determined.  In (b) the test data has been aligned to match the reference data at the center and a height offset is calculated.  Finally, in (c) the test data is adjusted by adding the height offset divided by two.
    }
    \label{fig:loc_alignment}
\end{figure}

This paragraph provides more details on the local alignment process.  The process is simply illustrated in Figure \ref{fig:loc_alignment} with smooth curves, but test data can have noise. First, the test data is modified so that the test data and reference data are aligned with respect to approximate minimum values over the center region. In particular, a zero-offset is determined by calculating the tenth percentile of the reference elevations over the center region minus the tenth percentile of the test elevations over the same region as in Figure \ref{fig:loc_alignment_a}. This zero-offset is added to the test data over the evaluation region. Next, a maximum elevation offset is calculated.  The ninetieth percentile of the test data is then calculated over the building evaluation regions, and a test maximum elevation is defined as the minimum of those two values.  The same calculation is done with the reference data to get a reference maximum elevation. A final adjustment to the test data is obtained by determining half of the difference in the reference and test maximum elevations. This offset is added to the test data and centers the test data between the bottom and top values of the reference data over the evaluation region. Figures \ref{fig:loc_alignment_b} and \ref{fig:loc_alignment_c} visualize the difference in max elevations and the result of adding half of that difference to obtain the final result.

\begin{figure}[h!]
    \centering
    \includegraphics[height=4.5cm]{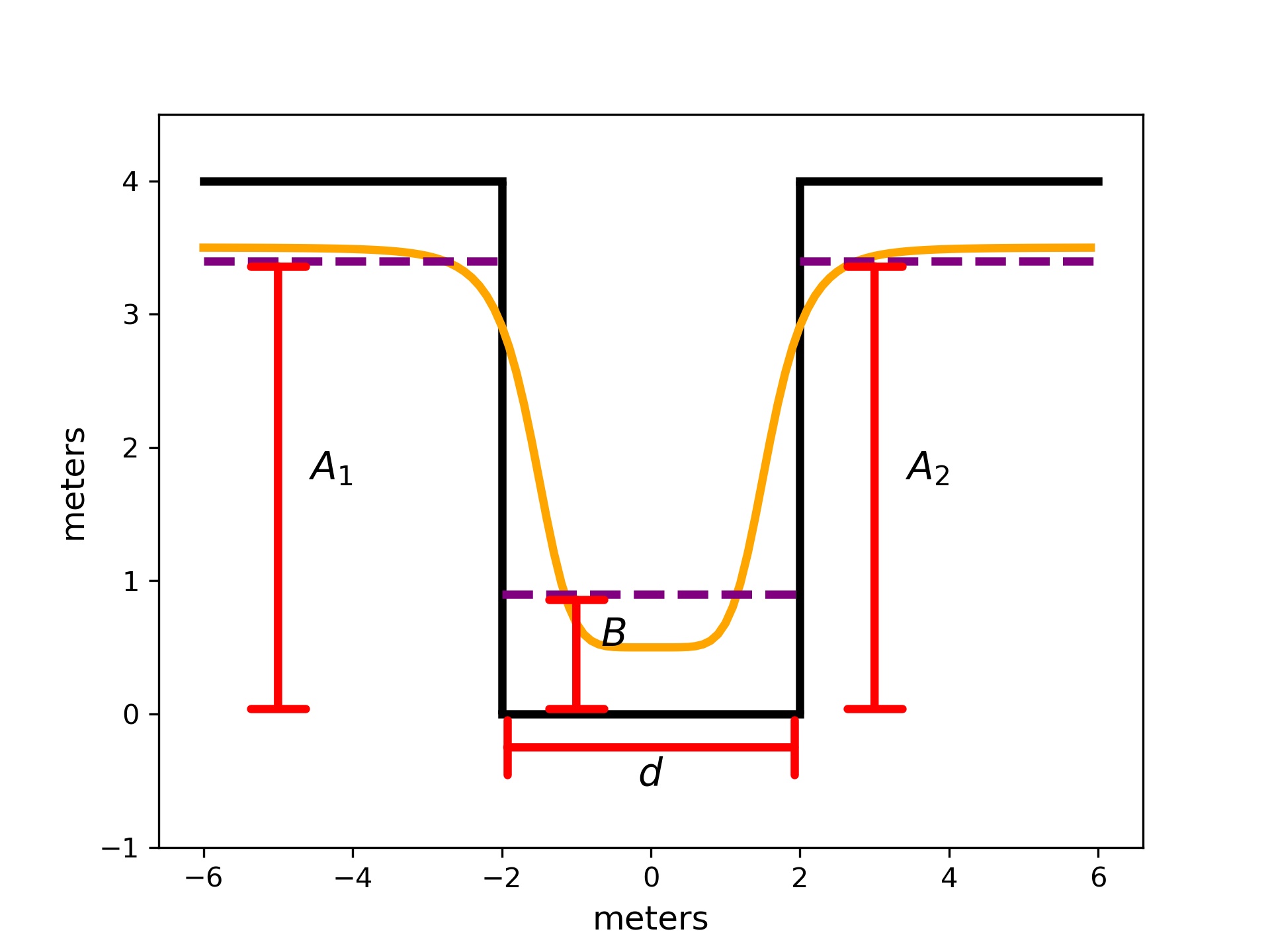} 
    \caption{Simple illustration of the calculation of CTF for an evaluation region. The black represents the reference data and the orange the locally aligned test data. The dashed purple lines used to measure $A_1$ and $A_2$ represent the two means of the clipped test data over the building evaluation regions. The dashed purple line used to measure $B_1$ is the mean of the clipped test data over the center region. (Clipping described in section \ref{sec:ctf_calc}.)}
    \label{fig:simple_ctf}
\end{figure}

The locally aligned test data is clipped to be between the tenth percentile of the reference data over the center region and the test maximum elevation.  The tenth percentile of the reference data is subtracted from the result so that the minimum value of the clipped data is zero.  The mean of the resulting test data is then calculated over the center and each building evaluation region.  Any data outside of $1.5$ times the inter-quartile range of the data is not included in the mean.  The CTF value reported for the building pair evaluation region is the mean of the two CTF values calculated; one for each building region with the same center region.  In particular, using the notation of the visualization in Figure \ref{fig:simple_ctf},
\begin{equation}
    C(d) = .5 \left(\frac{A_1 -B}{A_1+B} + \frac{A_2 - B}{A_2+B} \right).
\end{equation}

Note that issues may arise in using building pairs that do not occur when using a tribar measure.  For example, there may be obstructions in between the buildings such as trees.  There are filters in place to remove invalid results such as a reference CTF filter.  Only test CTF values for which the reference CTF is above a user determined threshold are included in the summary plot.  When using segmentation derived building footprints, it is possible that buildings were detected in reference data (transient structures) that were not present in the test data.  When this occurs, test CTF values of zero are to be obtained.  For true structures, CTF values of identically zero are not expected.  For this reason, we filter out CTF values that are identically zero. 

The function in Equation \eqref{eq:ctf_model} is fit to the CTF data and a horizontal line is plotted at the desired CTF threshold.  The intersection provides the distance at which the analyst may declare buildings are resolved.

\begin{figure}[ht]
    \begin{center}
        \begin{tabular}{cc} 
            \subcaptionbox{Reference CTF Values\label{fig:ctf_ref_legend}}{
                \includegraphics[height=4cm]{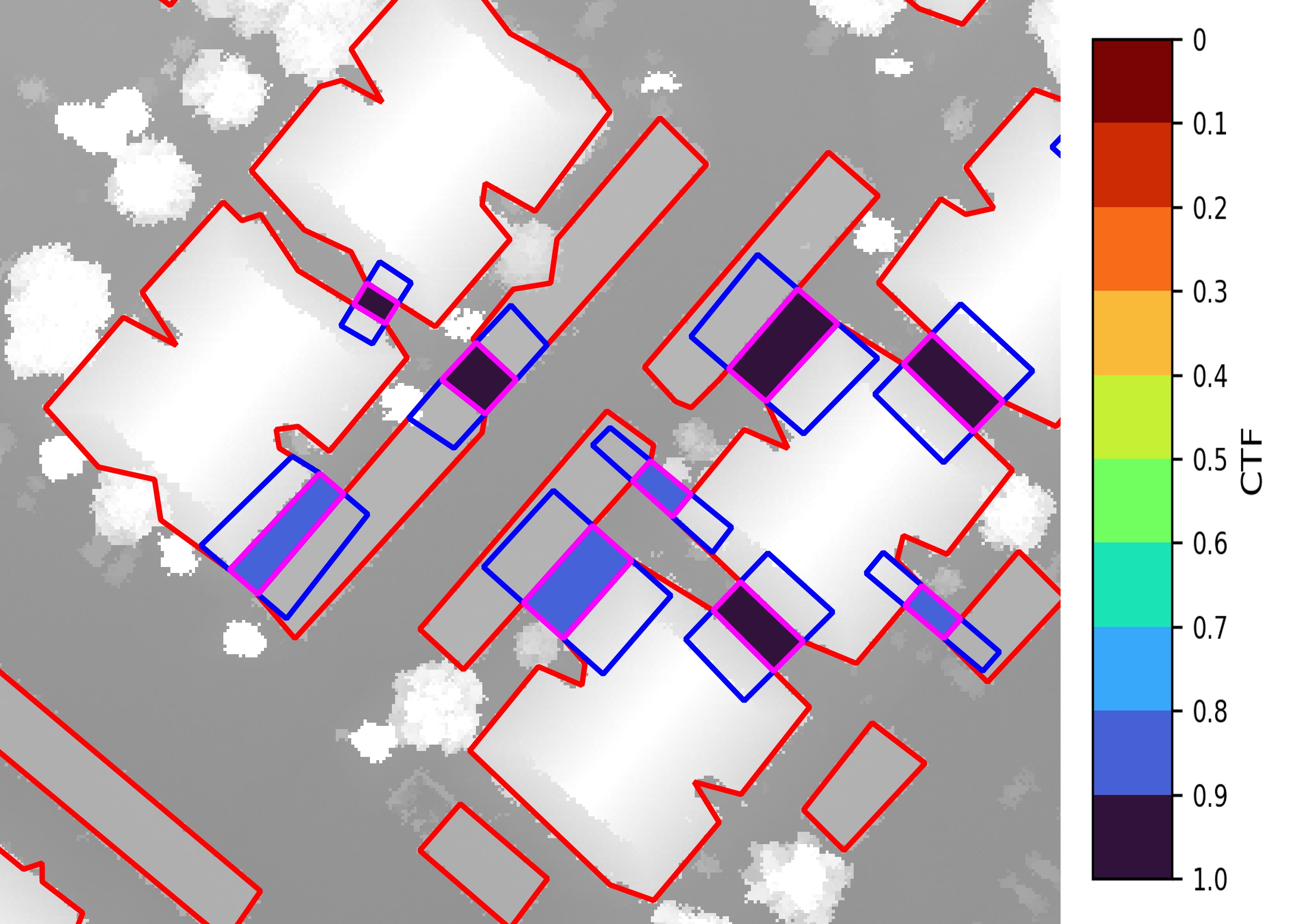}
            } &
            \subcaptionbox{Test CTF Values\label{fig:ctf_test_legend}}{
                \includegraphics[height=4cm]{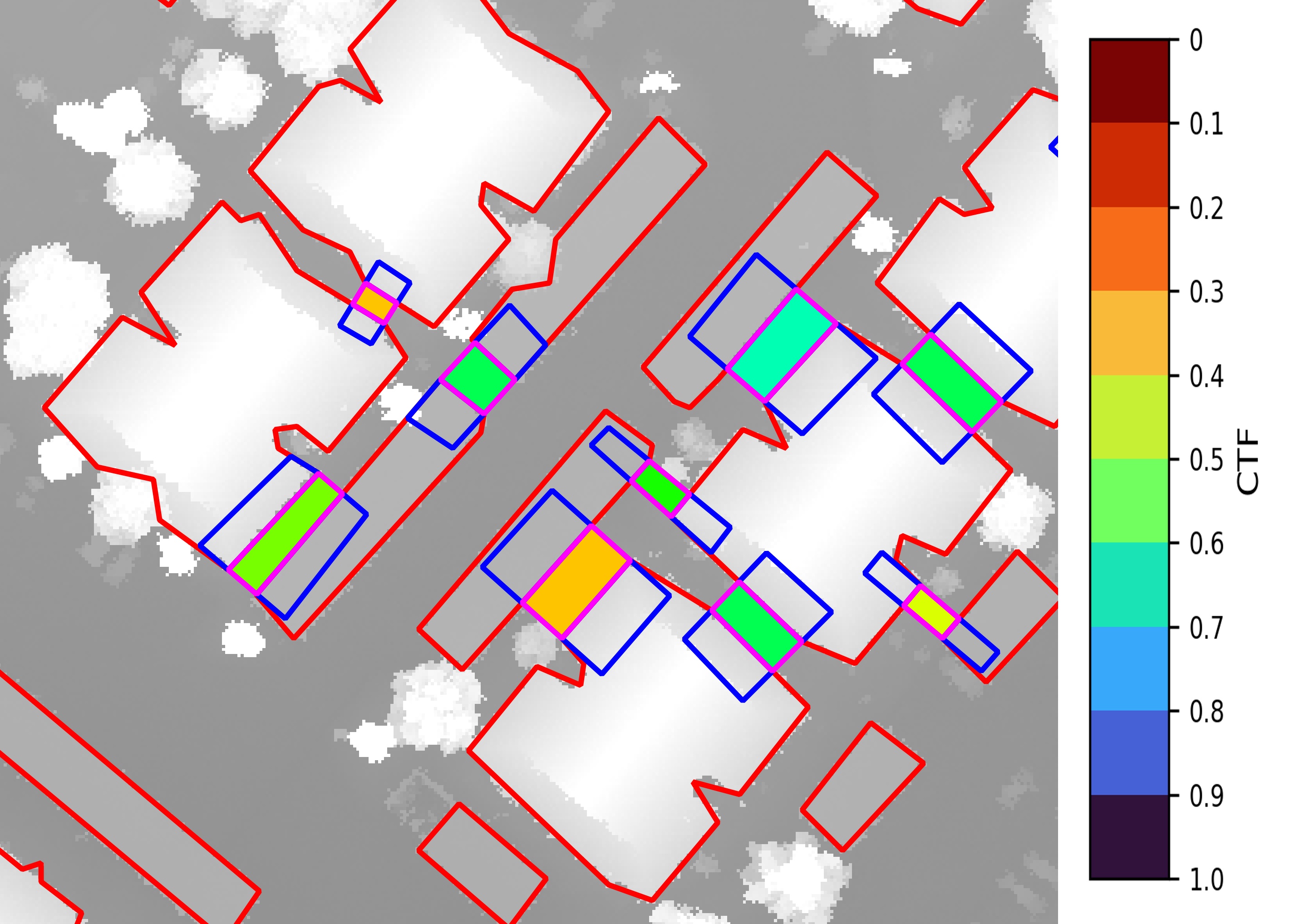}
            }
        \end{tabular}
    \end{center}
    \caption[example]{
        Comparison of CTF values over evaluation regions for reference and test data. Red denotes the lidar segmentation model derived footprints, blue the evaluation regions over the buildings and pink the center evaluation regions.  The colorbar indicates the calculated CTF values.  It is clear that the CTF values for the reference data are higher than the CTF values for the test data.  The  reference DSM was derived from the 3DEP lidar  point cloud \cite{USGS_3DEP} over a small region near Nellis AFB.
    }
    \label{fig:test_ref_ctf_compare} 
\end{figure}

\begin{figure}[ht]
    \begin{center}
        \begin{tabular}{ccc} 
            \subcaptionbox{Reference DSM\label{fig:bad_ref_region}}{
                \includegraphics[height=3.75cm]{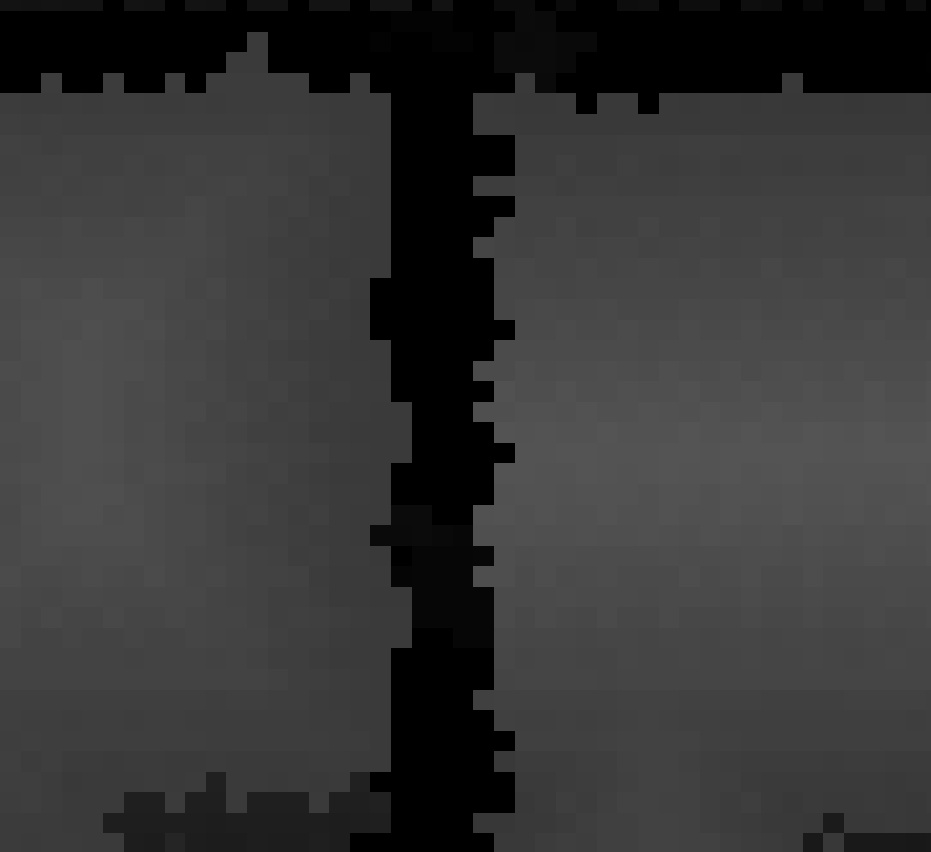}
            } &
            \subcaptionbox{Test DSM\label{fig:bad_test_region}}{
                \includegraphics[height=3.75cm]{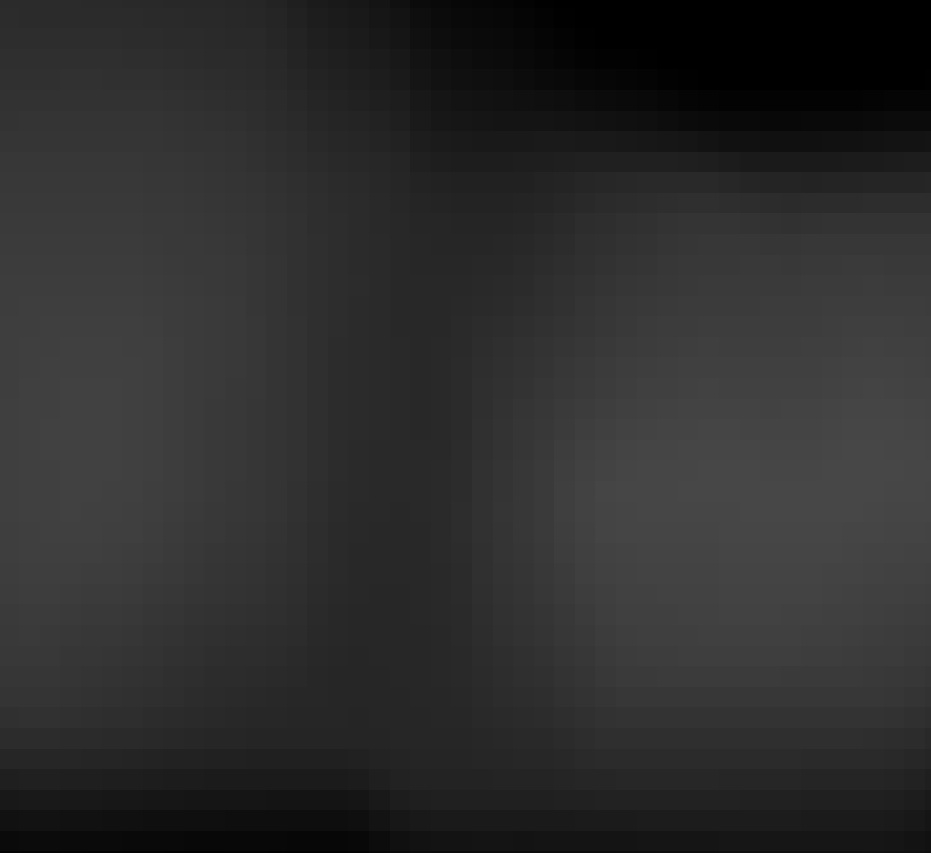}
            } &
            \subcaptionbox{Test CTF value\label{fig:bad_test_ctf_legend}}{
                \includegraphics[height=3.75cm]{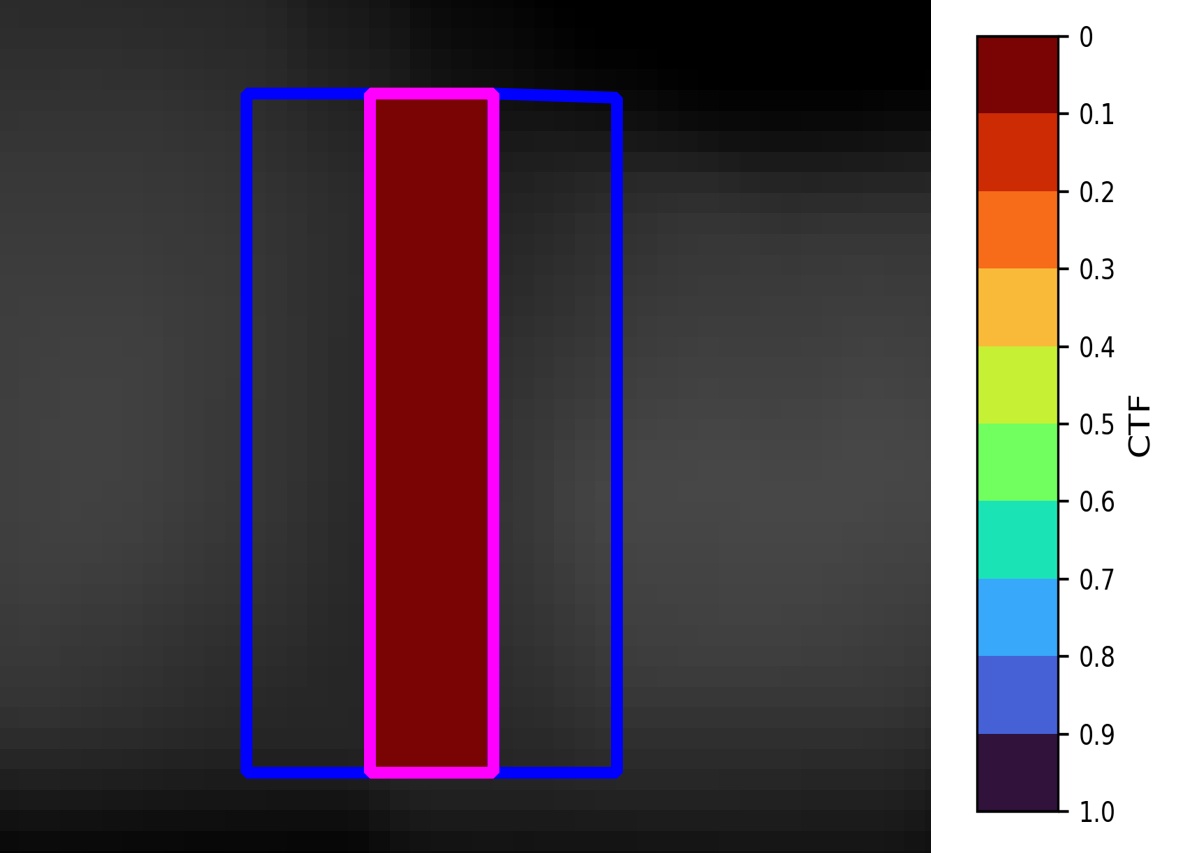}
            }
        \end{tabular}
    \end{center}
    \caption[example]{
       This illustration is an example of a CTF value below $.2$.  In (a) the building pair is clearly observed, while in the test data (b) the buildings are not easily distinguishable.  In (c) the building evaluation regions are outlined in blue and the center evaluation region in pink. The reference DSM was derived from the 3DEP lidar  point cloud \cite{USGS_3DEP} over a small region near Nellis AFB and the test DSM was derived from MAXAR\textsuperscript{\copyright} satellite imagery, USGP-20211007.
    }
    \label{fig:bad_ctf}
\end{figure}

\begin{figure}[ht]
    \begin{center}
        \begin{tabular}{cc} 
            \subcaptionbox{Lidar (reference CTF $> .95$)\label{fig:lid_95}}{
                \includegraphics[height=4.2cm]{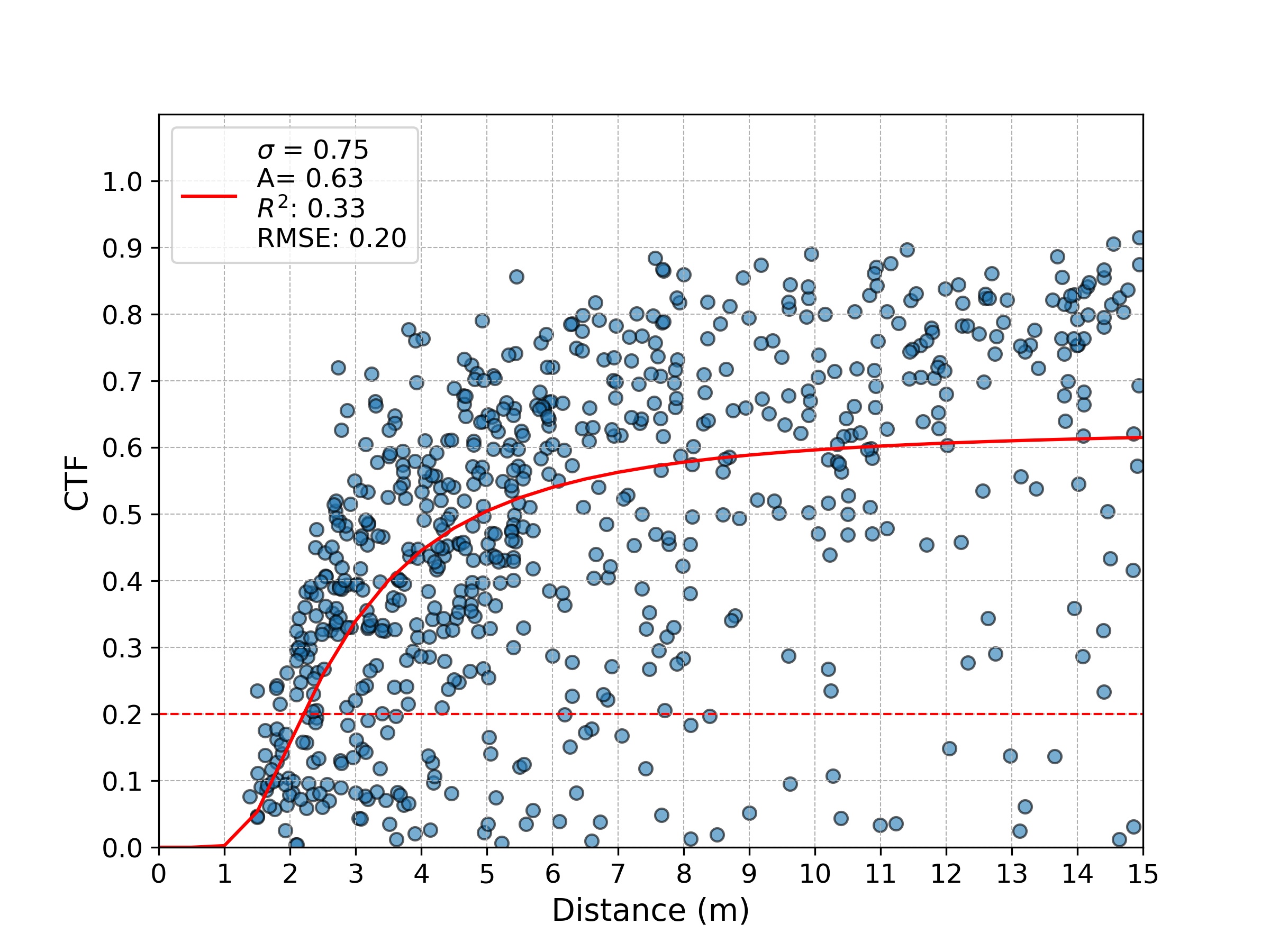}
            } &
            \subcaptionbox{Lidar (reference CTF $> .98$)\label{fig:lid_98}}{
                \includegraphics[height=4.2cm]{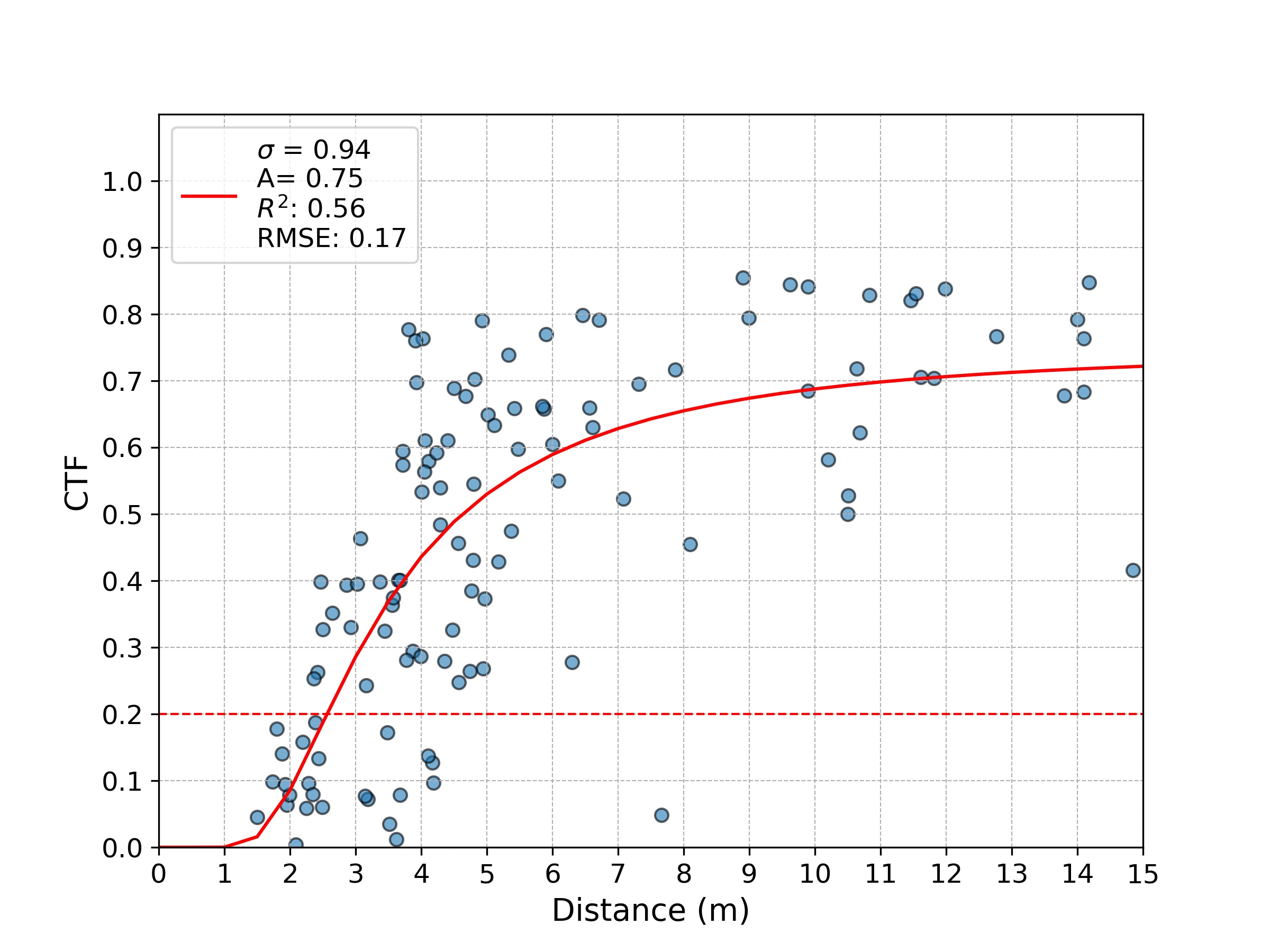}
            } \\\\
            \multicolumn{2}{c}{
                \subcaptionbox{Aligned OSM (reference CTF $>.95$)\label{fig:osm_95}}{
                    \includegraphics[height=4.2cm]{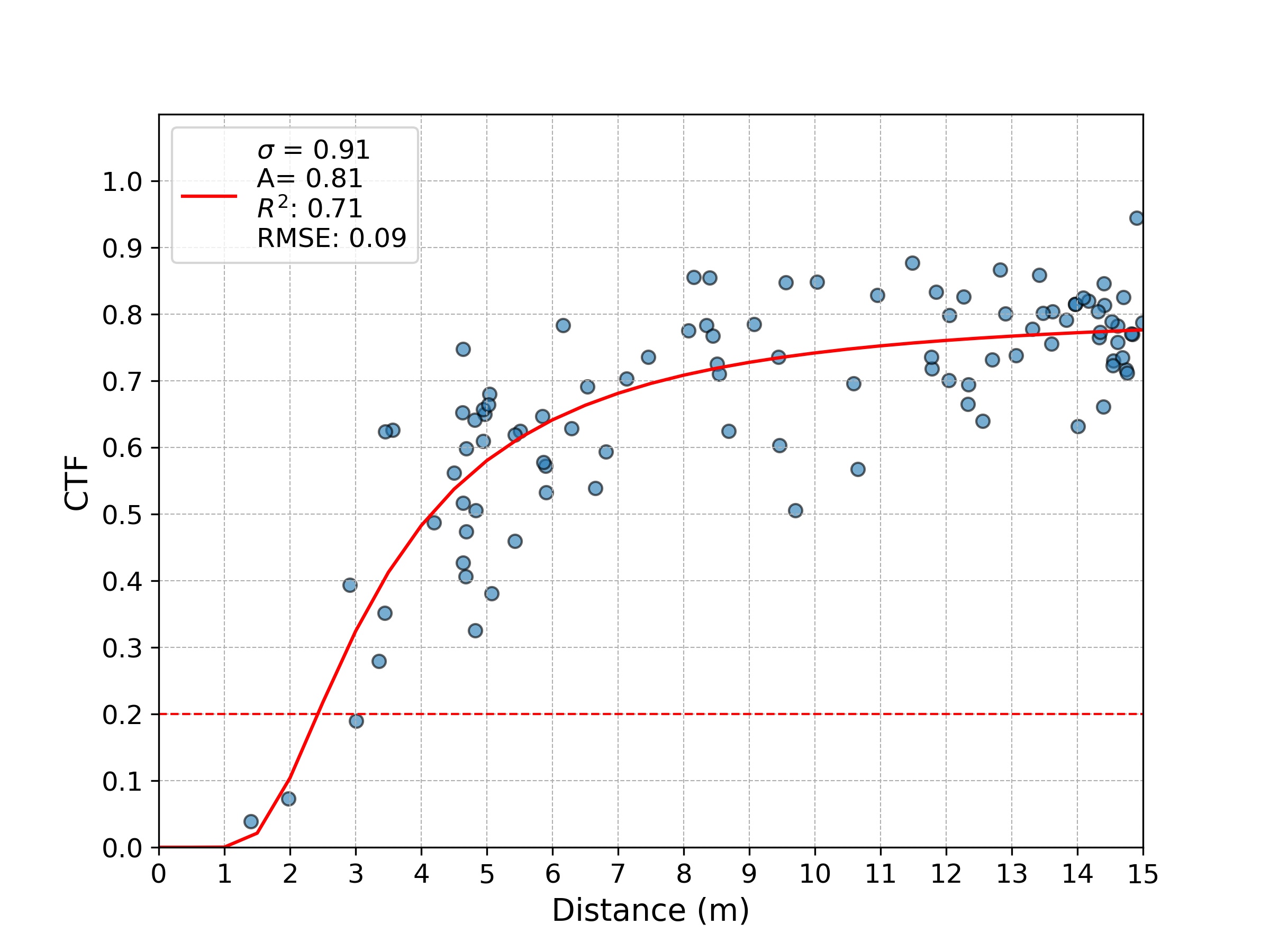}
                }
            }
        \end{tabular}
    \end{center}
    \caption{Example CTF summary plots over Nellis. Results in (a) and (b) were obtained using evaluation regions obtained from lidar segmentation model derived building footprints and show the impact of filtering with respect to the reference CTF values. The result using evaluation pairs obtained from aligned OSM building footprints is in (c). The distance of approximately $d=2.5\, \mathrm{m}$ at CTF$=.2$ closely agrees in all three examples.}
    \label{fig:summary_plot}
\end{figure}

The visualization of corresponding CTF values for the previous examples are in Figure \ref{fig:test_ref_ctf_compare}.  An example of a low ($<.2$) test CTF score is provided in Figure \ref{fig:bad_ctf}.  Summary plots for the entire region are provided in Figure \ref{fig:summary_plot}, including summary plots using lidar segmentation model derived results with differing reference CTF filter thresholds and OSM derived results.

\subsection{Considerations}
For a successful evaluation it is necessary to have a good distribution of buildings with a variety of distances between them that are not obstructed. It has been observed that the segmentation model does not perform as well for large buildings, so possible evaluation pairs may be missed if a region has only very large buildings.

Differences in the target and reference data (building changes) will create low or zero CTF values if a building present in the reference data is not present in the test data.  These are currently filtered out of the analysis.  

There are several reasons why the CTF data will have significant variance.  For example, elevation changes in the region, the relationship between a location and the viewing angle geometry of input satellite imagery, misalignment in building footprints (when not using lidar), and inaccurate lidar-derived building footprints.  

Finally, keep in mind the notes about the interpretation of the analysis found at the end of section \ref{sec:ctf}.

\subsection{Other Metrics}
With curated reference data and aligned test data established in our evaluation pipeline, other metrics can be calculated as well. Similar to Bosch et al.\cite{isprs-archives-XLII-1-W1-239-2017} and Hagstrom et al. \cite{9554754}, RMSE and percentile statistics can be computed for relative vertical accuracy and visualizations produced to highlight any low accuracy features.

\section{ADDITIONAL RESULTS}
\begin{figure}[ht]
    \begin{center}
        \begin{tabular}{cc}
            \subcaptionbox{Jacksonville Imagery \label{fig:jack_im}}{
                \includegraphics[height=4cm]{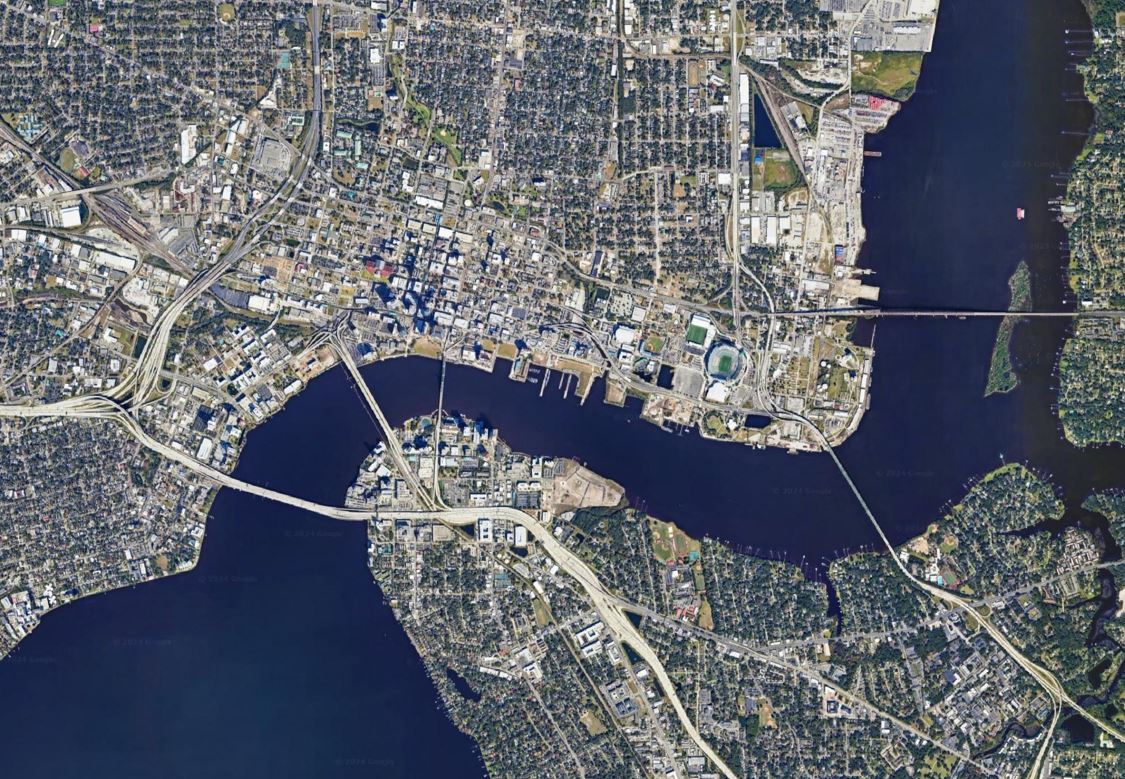}
            } &
            \subcaptionbox{Jacksonville Test Data\label{fig:jack_test}}{
                \includegraphics[height=4cm]{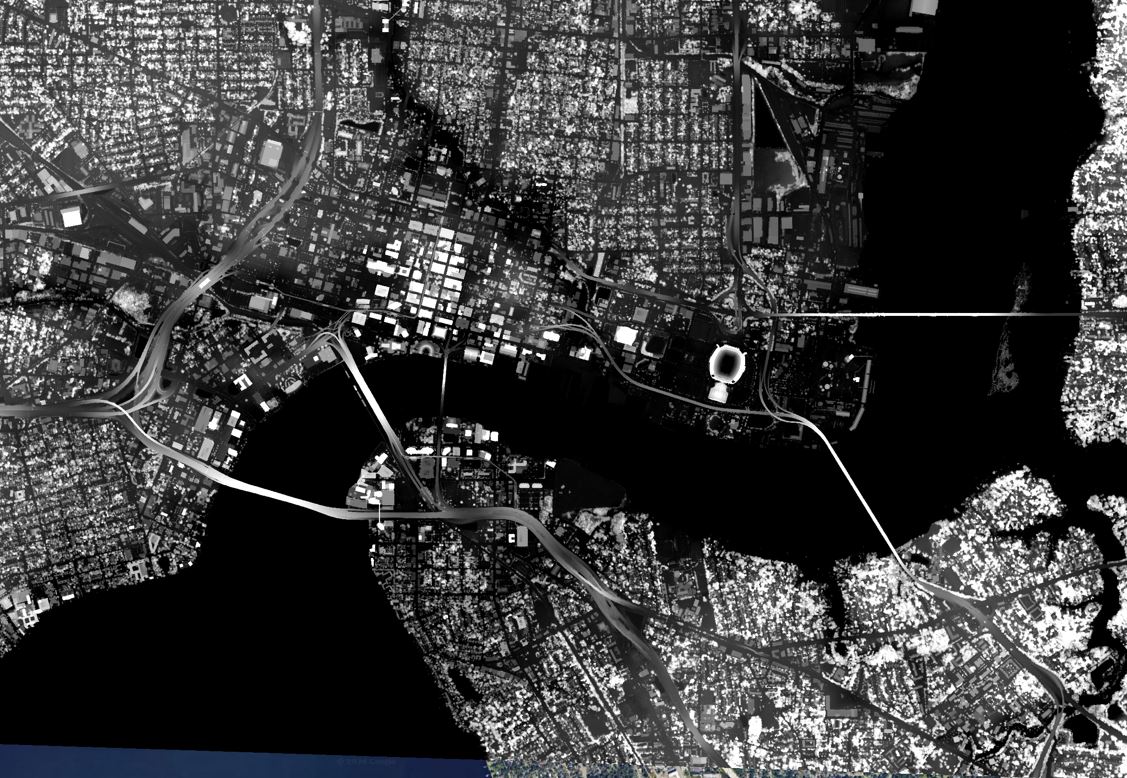}
            } 
        \end{tabular}
    \end{center}
    \caption[example]{
        Google Earth visualization, Google\textsuperscript{\copyright} 2023 (a) of approximately 50 square kilometer region over Jacksonville, Florida for which example CTF example analysis has been completed for test DSM shown in (b).  The test DSM was derived from MAXAR\textsuperscript{\copyright} satellite imagery, USGP-20211007.
    }
    \label{fig:jack_data}
\end{figure}

\begin{figure}[ht]
    \begin{center}
        \begin{tabular}{cc} 
            \subcaptionbox{Lidar (reference $> .95$)\label{fig:ctf_jack_lid}}{
                \includegraphics[height=4.5cm]{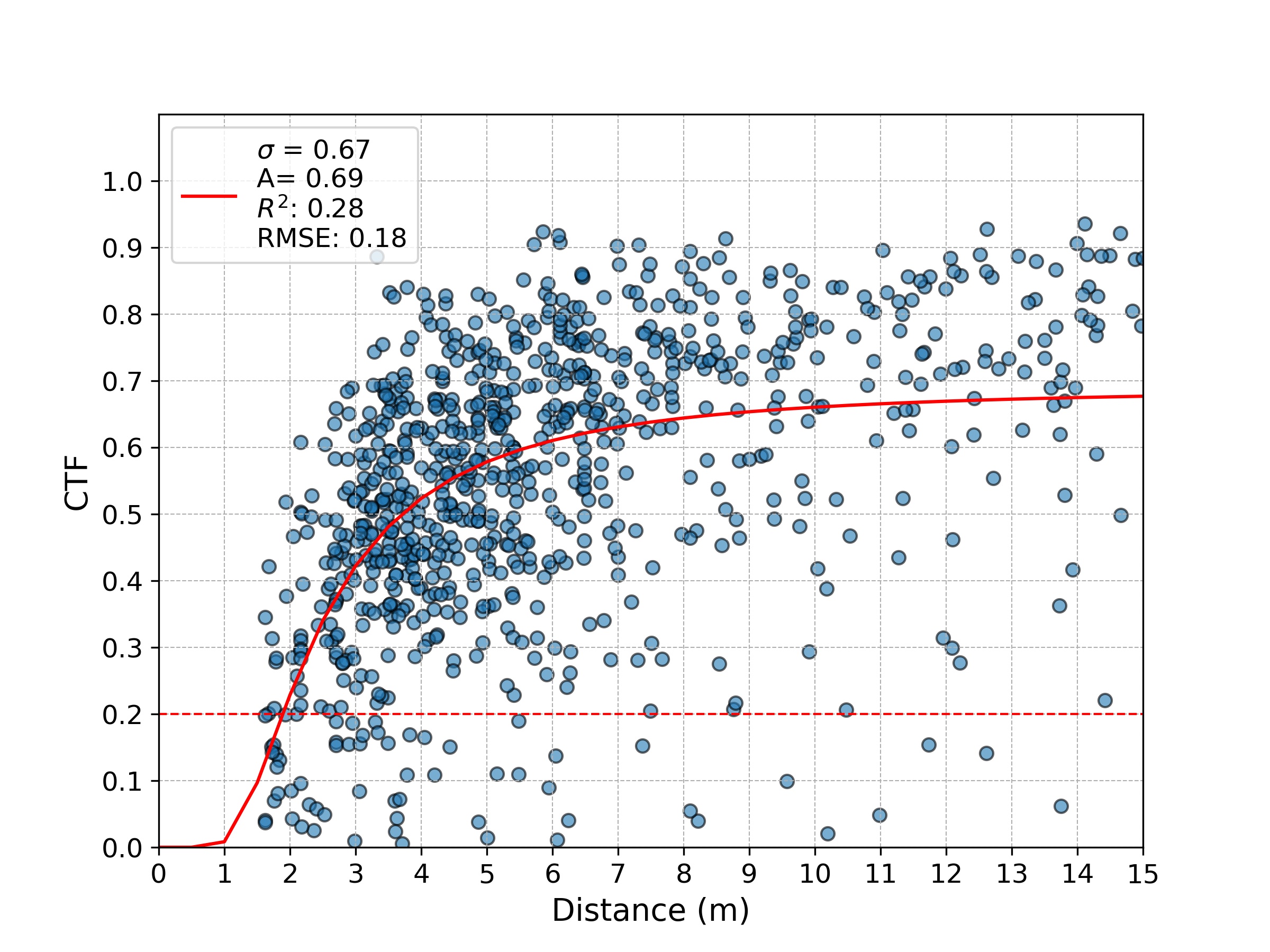}
            } &
            \subcaptionbox{Hand-Curated (reference $>.95)$\label{fig:ctf_jack_hc}}{
                \includegraphics[height=4.5cm]{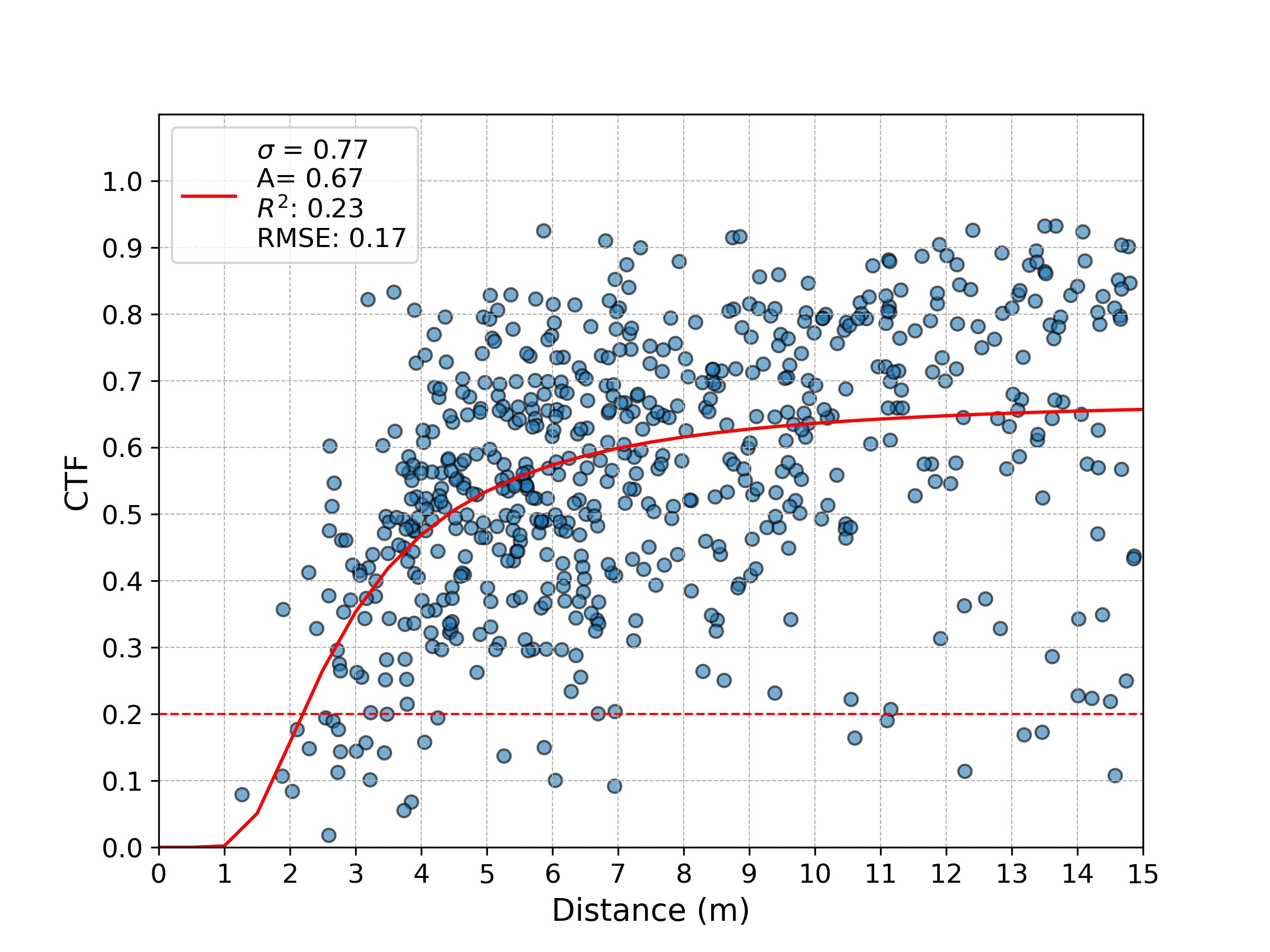}
            }
        \end{tabular}
    \end{center}
    \caption[example]{
        CTF summary plots are provided with respect to the evaluation regions displayed in Figure \ref{fig:ctf_jack_locs}.  The distance value at CTF$=.2$ is approximately $2\, \mathrm{m}$ in both plots.  It is slightly larger in the hand-curated CTF summary but there are fewer points with small distance.
    }
    \label{fig:jack_ctf_summaries} 
\end{figure}

\begin{figure}[ht]
    \begin{center}
        \begin{tabular}{cc}
            \subcaptionbox{Lidar Segmentation\label{fig:eval_lidar}}{
                \includegraphics[height=4cm]{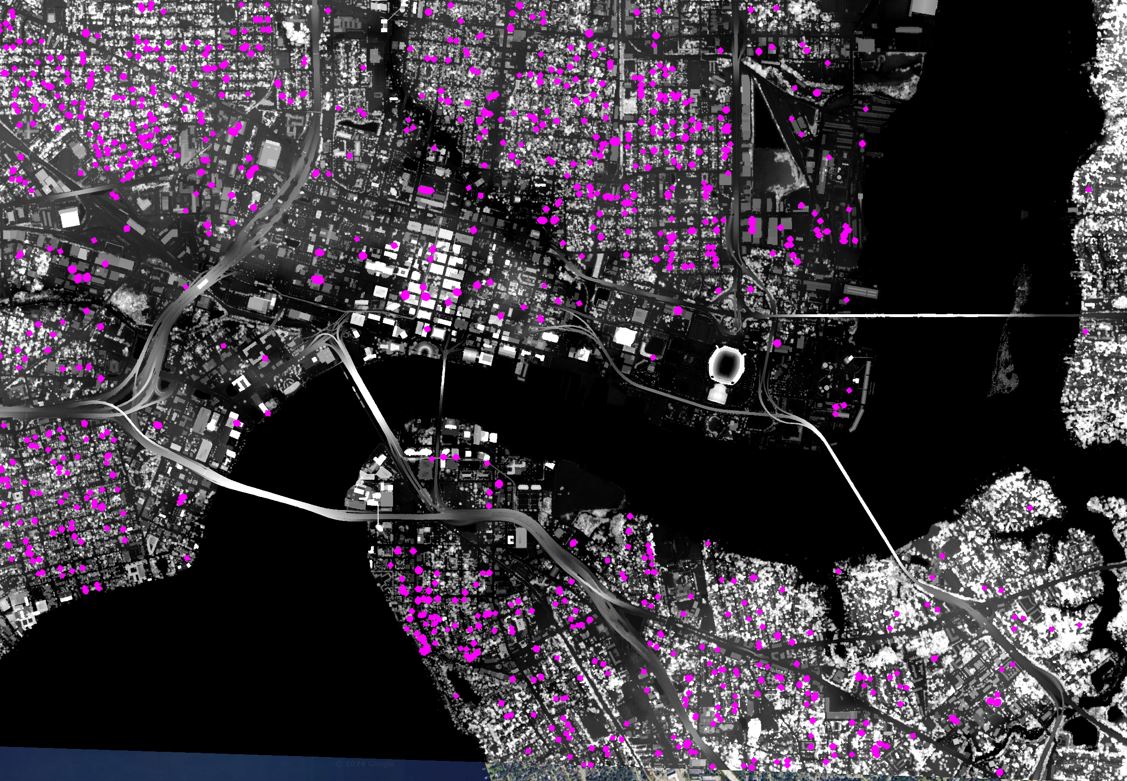}
            } &
            \subcaptionbox{Hand-Curated\label{fig:eval_hc}}{
                \includegraphics[height=4cm]{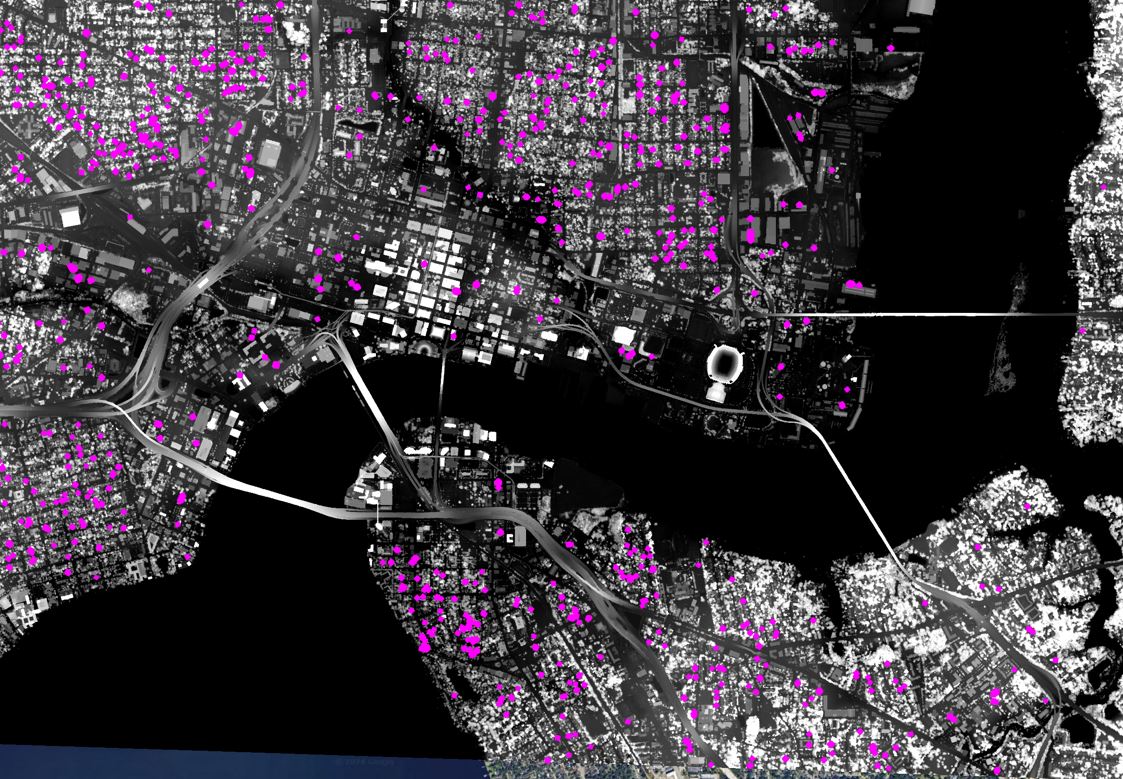}
            } 
        \end{tabular}
    \end{center}
    \caption[example]{
        The evaluation region centers over a reference DSM derived from 3DEP lidar data \cite{USGS_3DEP} over Jacksonville are visualized in pink with respect to the lidar segmentation model derived footprints (a) and the example hand-curate building footprints (b).  The distribution of the evaluation regions over the region is quite good in this example; they are not clustered in a few subregions.
    }
    \label{fig:ctf_jack_locs}
\end{figure}

CTF results from another region are shared in this section to demonstrate the utility of the methodology.  An approximately $50$ square kilometer region over Jacksonville, Florida is shown in Figure \ref{fig:jack_data} along with a corresponding test DSM.  The CTF results for the region with respect to lidar segmentation model derived footprints and example provided footprints are shown in Figure \ref{fig:jack_ctf_summaries}.  The location of the corresponding evaluation regions for the plots are shown in Figure \ref{fig:ctf_jack_locs}.  The region is a good candidate for CTF analysis because there are many buildings with differing distances between them that cover the entire region fairly consistently.

\section{CONCLUSIONS}
This paper has introduced an adaptation of the contrast transfer function (CTF) for evaluating resolution of a provided 3D data product using reference lidar data.  The technique requires the presence of parallel buildings with little to no obstruction between them and a variety of distances between them.  The technique allows an analyst to compare resolutions of multiple methodologies for producing a 3D data product.  The CTF pipeline produces aligned test and reference data that can be compared at a pixel level.  This enables the calculation of other metrics for future studies.

Future work includes continued improvement to the lidar segmentation model and more advanced techniques for obtaining building footprint polygons.  Improvements can also be made in the determination of evaluation regions so that more data points are available for analysis. For test products obtained from photogrammetric techniques, it would be interesting to study the relationship between CTF and metadata for the input imagery.  The presented techniques could also potentially be altered to study the size of a minimal discernible object for a given test DSM.

\acknowledgments 
This material is based upon work supported by the National Geospatial-Intelligence Agency (NGA) under Contract No. HM047623D0003 HM04762323F0105. Any opinions, findings and conclusions or recommendations expressed in this material are those of the author(s) and do not necessarily reflect the views of NGA, DoD, or the US government. Approved for public release, NGA-U-2025-00517.


\begin{thebibliography}{10}

\bibitem{rsp_pipeline}
Qin, R., ``{RPC STEREO PROCESSOR (RSP) – A SOFTWARE PACKAGE FOR DIGITAL SURFACE MODEL AND ORTHOPHOTO GENERATION FROM SATELLITE STEREO IMAGERY},'' {\em ISPRS Annals of Photogrammetry, Remote Sensing and Spatial Information Sciences}~{\bf III-1} (2016).

\bibitem{s2p8014932}
{Facciolo}, G., {De Franchis}, C., and {Meinhardt-Llopis}, E., ``Automatic 3d reconstruction from multi-date satellite images,'' {\em CVPRW}  (2017).

\bibitem{VisSat19}
Zhang, K., Sun, J., and Snavely, N., ``{Leveraging Vision Reconstruction Pipelines for Satellite Imagery},'' {\em ICCV Workshop on 3D Reconstruction in the Wild}  (2019).

\bibitem{Leotta_2019_CVPR_Workshops}
Leotta, M.~J., Long, C., Jacquet, B., Zins, M., Lipsa, D., Shan, J., Xu, B., Li, Z., Zhang, X., Chang, S.-F., Purri, M., Xue, J., and Dana, K., ``Urban semantic 3d reconstruction from multiview satellite imagery,'' {\em CVPRW}  (2019).

\bibitem{mari2022sat}
Mar{\'\i}, R., Facciolo, G., and Ehret, T., ``{Sat-NeRF}: Learning multi-view satellite photogrammetry with transient objects and shadow modeling using {RPC} cameras,'' {\em 2022 IEEE/CVF Conference on Computer Vision and Pattern Recognition Workshops}  (2022).

\bibitem{Mari_2023_CVPR}
Mar{\'\i}, R., Facciolo, G., and Ehret, T., ``Multi-date earth observation nerf: The detail is in the shadows,'' {\em Proceedings of the IEEE/CVF Conference on Computer Vision and Pattern Recognition Workshops}  (2023).

\bibitem{zhang2024brdfnerfneuralradiancefields}
Zhang, L., Rupnik, E., Nguyen, T.~D., Jacquemoud, S., and Klinger, Y., ``{BRDF-NeRF: Neural Radiance Fields with Optical Satellite Images and BRDF Modelling},'' {\em arXiv}  (2024).

\bibitem{sprintson2024fusionrfhighfidelitysatelliteneural}
Sprintson, M., Chellappa, R., and Peng, C., ``{FusionRF: High-Fidelity Satellite Neural Radiance Fields from Multispectral and Panchromatic Acquisitions},'' {\em arXiv}  (2024).

\bibitem{isprs-archives-XLII-1-W1-239-2017}
Bosch, M., Leichtman, A., Chilcott, D., Goldberg, H., and Brown, M., ``{METRIC EVALUATION PIPELINE FOR 3D MODELING OF URBAN SCENES},'' {\em The International Archives of the Photogrammetry, Remote Sensing and Spatial Information Sciences}~{\bf XLII-1/W1} (2017).

\bibitem{Wang_2023_ICCV}
Wang, R., Huang, S., and Yang, H., ``{Building3D: A Urban-Scale Dataset and Benchmarks for Learning Roof Structures from Point Clouds},'' {\em Proceedings of the IEEE/CVF International Conference on Computer Vision}  (2023).

\bibitem{2011_Stevens}
Stevens, J.~R., Lopez, N.~A., and Burton, R.~R., ``Quantitative data quality metrics for 3d laser radar systems,'' {\em Laser Radar Technology and Applications XVI}  (2011).

\bibitem{schott2007remote}
Schott, J., {\em {Remote Sensing: The Image Chain Approach}}, Oxford University Press (2007).

\bibitem{OpenStreetMap}
{OpenStreetMap contributors}, ``{Planet dump retrieved from https://planet.osm.org}.'' \url{https://www.openstreetmap.org} (2017).

\bibitem{USGS_3DEP}
{U.S. Geological Survey}, ``3d elevation program (3dep) lidar data.'' The National Map (2021).
\newblock Accessed March 15, 2024.

\bibitem{liu2022spunetselfsupervisedpointcloud}
Liu, X., Liu, X., Liu, Y.-S., and Han, Z., ``{SPU-Net: Self-Supervised Point Cloud Upsampling by Coarse-to-Fine Reconstruction With Self-Projection Optimization},'' {\em Trans. Img. Proc.}~{\bf 31} (2022).

\bibitem{pointcept2023}
{Pointcept Contributors}, ``{Pointcept: A Codebase for Point Cloud Perception Research}.'' \url{https://github.com/Pointcept/Pointcept} (2023).

\bibitem{varney2020dales}
Varney, N., Asari, V.~K., and Graehling, Q., ``{DALES: A Large-scale Aerial LiDAR Data Set for Semantic Segmentation},'' {\em Proceedings of the IEEE/CVF Conference on Computer Vision and Pattern Recognition Workshops}  (2020).

\bibitem{Chen_2022_BMVC}
Chen, M., Hu, Q., Yu, Z., THOMAS, H., Feng, A., Hou, Y., McCullough, K., Ren, F., and Soibelman, L., ``{STPLS3D: A Large-Scale Synthetic and Real Aerial Photogrammetry 3D Point Cloud Dataset},'' {\em 33rd British Machine Vision Conference}  (2022).

\bibitem{swissSURFACE3D}
{Federal Office of Topography swisstopo}, ``{swissSURFACE3D}.'' \url{https://www.swisstopo.admin.ch/en/height-model-swisssurface3d}.
\newblock Accessed: 2025-01-21.

\bibitem{douglas1973algorithms}
Douglas, D.~H. and Peucker, T.~K., ``Algorithms for the reduction of the number of points required to represent a digitized line or its caricature,'' {\em Cartographica: The International Journal for Geographic Information and Geovisualization}~{\bf 10} (1973).

\bibitem{Girard_2021_CVPR}
Girard, N., Smirnov, D., Solomon, J., and Tarabalka, Y., ``Polygonal building extraction by frame field learning,'' {\em Proceedings of the IEEE/CVF Conference on Computer Vision and Pattern Recognition}  (2021).

\bibitem{9554754}
Hagstrom, S., Pak, H.~W., Ku, S., Wang, S., Hager, G., and Brown, M., ``Cumulative assessment for urban 3d modeling,'' {\em 2021 IEEE International Geoscience and Remote Sensing Symposium}  (2021).

\end{thebibliography}

\end{document}